%% file: main.tex
\documentclass{article}
\pdfoutput=1
\PassOptionsToPackage{numbers, compress}{natbib}

\usepackage[final]{neurips_2024}

\usepackage[utf8]{inputenc} 
\usepackage[T1]{fontenc}    
\usepackage{hyperref}       
\usepackage{url}            
\usepackage{booktabs}       
\usepackage{amsfonts}       
\usepackage{nicefrac}       
\usepackage{microtype}      
\usepackage{xcolor}         
\usepackage{amsmath}

\usepackage{algorithm}  
\usepackage{algorithmicx}  
\usepackage{algpseudocode}  

\usepackage{wrapfig}
\usepackage{adjustbox}

\usepackage{enumitem}
\usepackage{graphicx}
\usepackage{subfigure}
\usepackage{bm}
\usepackage{comment}
\usepackage{CJKutf8}

\usepackage{subfigure}

\definecolor{MyDarkBlue}{rgb}{0,0.5,1}
\definecolor{MyDarkGreen}{rgb}{0.02,0.6,0.02}
\definecolor{MyDarkRed}{rgb}{0.8,0.02,0.02}
\definecolor{MyDarkOrange}{rgb}{0.40,0.2,0.02}
\definecolor{MyYellow}{rgb}{1,0.55,0}
\definecolor{MyPurple}{RGB}{111,0,255}
\definecolor{MyRed}{rgb}{1.0,0.0,0.0}
\definecolor{MyGold}{rgb}{0.75,0.6,0.12}
\definecolor{MyDarkgray}{rgb}{0.66, 0.66, 0.66}
\definecolor{default}{RGB}{0,0,0}

\newcommand\ie{\textit{i.e., }}
\newcommand{\model}{CoWorld} 
\renewcommand{\eqref}[1]{Eq.~(\ref{#1})} %
\newcommand{\figref}[1]{Figure~\ref{#1}} %
\newcommand{\tabref}[1]{Table~\ref{#1}} %

\newcommand{\padspace}{\hspace{3.5em}}

\title{Making Offline RL Online: Collaborative World Models for Offline Visual Reinforcement Learning}

\author{
Qi Wang$^{1,2}$\thanks{Equal contribution.} 
\ \
Junming Yang$^{3*}$ 
\ \
Yunbo Wang$^{1}$\thanks{Corresponding author: Yunbo~Wang~<yunbow@sjtu.edu.cn>.}
\ \
Xin Jin$^{2}$
\ \
Wenjun Zeng$^{2}$
\ \
Xiaokang Yang$^{1}$\\
$^1$ MoE Key Lab of Artificial Intelligence, AI Institute, Shanghai Jiao Tong University, China\\
$^2$ Ningbo Institute of Digital Twin, Eastern Institute of Technology, China\\
$^3$ School of Computer Science and Engineering, Southeast University, China \\
\textcolor{magenta}{\url{https://qiwang067.github.io/coworld}}
}

\begin{document}

\maketitle

\input{sections/00.abstract}
\input{sections/01.intro}

\input{sections/02.problem_setup}

\input{sections/03.method}

\input{sections/04.experiments}
\input{sections/05.related}
\input{sections/06.conclusion}


\medskip
{
\small

\input{main.bbl}

}


\input{sections/07.appendix}


\end{document}

%% file: sections/00.abstract.tex
\vspace{-5pt}
\begin{abstract}

Training offline RL models using visual inputs poses two significant challenges, \textit{i.e.}, the overfitting problem in representation learning and the overestimation bias for expected future rewards. Recent work has attempted to alleviate the overestimation bias by encouraging conservative behaviors. This paper, in contrast, tries to build more flexible constraints for value estimation without impeding the exploration of potential advantages. The key idea is to leverage off-the-shelf RL simulators, which can be easily interacted with in an online manner, as the ``\textit{test bed}'' for offline policies. To enable effective online-to-offline knowledge transfer, we introduce CoWorld, a model-based RL approach that mitigates cross-domain discrepancies in state and reward spaces. Experimental results demonstrate the effectiveness of CoWorld, outperforming existing RL approaches by large margins.

\end{abstract}

%% file: sections/01.intro.tex
\vspace{-5pt}
\section{Introduction}
\vspace{-3pt}
Learning control policies with visual observations can be challenging due to high interaction costs with the physical world.
Offline reinforcement learning (RL) is a promising approach to address this challenge~\citep{fujimoto2019off,kumar2020conservative,qi2022data,chen2023offline,zhuang2023behavior}.
However, the direct use of current offline RL algorithms in visual control tasks presents two primary difficulties. 
Initially, \textit{offline visual RL} is more prone to overfitting issues during representation learning, as it involves extracting hidden states from the limited, high-dimensional visual inputs.
Moreover, like its state-space counterpart, offline visual RL is susceptible to the challenge of value overestimation, as we observe from existing methods~\cite{laskin2020curl,hafner2021mastering}.

\begin{wrapfigure}{r}{0.52\textwidth}
\vspace{-12pt}
\includegraphics[width=\linewidth]{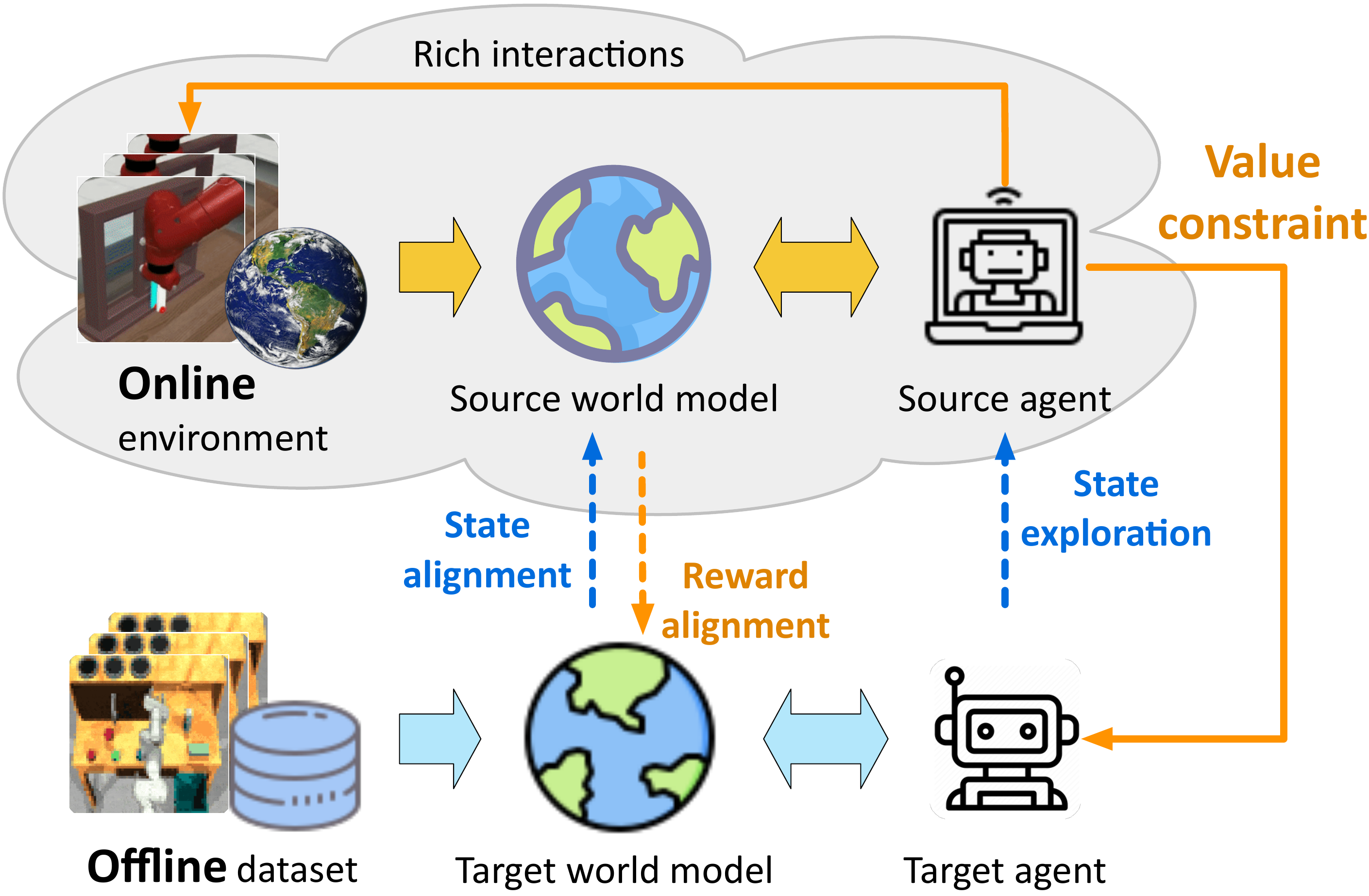}
\vspace{-15pt}
\caption{Our approach for offline visual RL.
}
\label{fig:key_ins}
\end{wrapfigure}

Improving offline visual RL remains an under-explored research area.
We aim to balance between overestimating and over-conservatism of the value function to avoid excessively penalizing the estimated values beyond the offline data distribution.
Intuitively, \textbf{\emph{we should not overly constrain the exploration with potential advantages.}}
Our basic idea, as illustrated in \figref{fig:key_ins}, is to leverage readily available online simulators for related (not necessarily identical) visual control tasks as auxiliary source domains, so that we can frame offline visual RL as an \textit{offline-online-offline} transfer learning problem to learn mildly conservative policies.

We present a novel model-based transfer RL approach called Collaborative World Models (\model{}).
Specifically, we train separate world models and RL agents for source and target domains, each with domain-specific parameters. To mitigate discrepancies between the world models, we introduce a novel representation learning scheme comprising two iterative training stages. These stages, as shown in \figref{fig:key_ins}, facilitate the alignment of latent state distributions (\textit{offline to online}) and reward functions (\textit{online to offline}), respectively.
By doing so, the source domain critic can serve as an online ``test bed'' for assessing the target offline policy. It is also more ``knowledgeable'' as it can actively interact with the online environment and gather rich information. 
Another benefit of the domain-collaborative world models is the ability to alleviate overfitting issues of offline representation learning, leading to more generalizable latent states derived from limited offline visual data.

\begin{figure*}[t]
    \centering
\includegraphics[width=\linewidth]{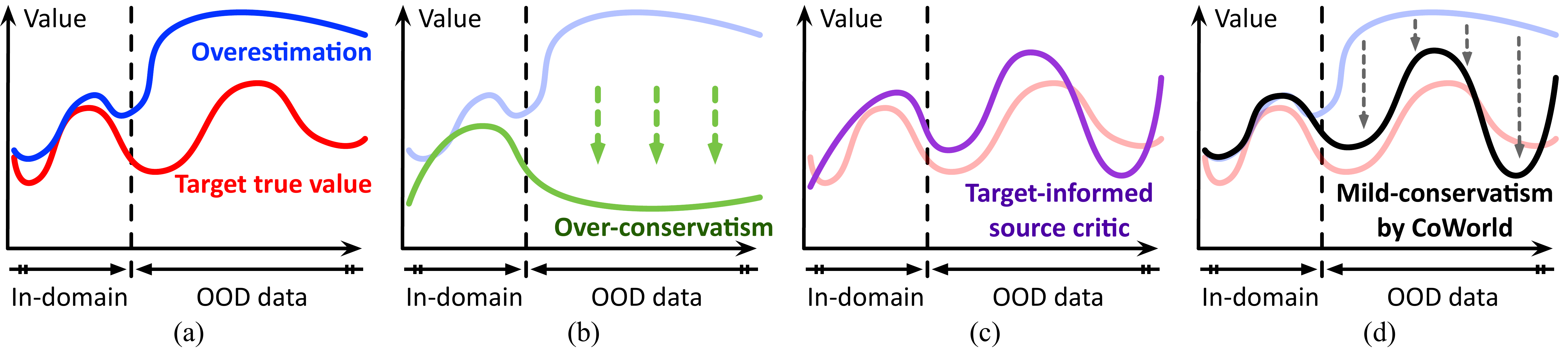}
  \vspace{-15pt}
    \caption{To address value overestimation in offline RL \textbf{(a)}, we can directly penalize the estimated values beyond the distribution of offline data, which may hinder the agent's exploration of potential states with high rewards \textbf{(b)}. Unlike existing methods, \model{} trains a cross-domain critic model in an online auxiliary domain to reassess the offline policy \textbf{(c)}, and regularizes the target values with flexible constraints \textbf{(d)}. The feasibility of this approach lies in the domain alignment techniques during the world model learning stage.
    }
    \label{fig:intro2}
    \vspace{-10pt}
\end{figure*}

For behavior learning in the offline dataset, we exploit the knowledge from the source model and introduce a mild regularization term to the training objective of the target domain critic model.
This regularization term encourages the \textbf{source critic} to reevaluate the \textbf{target policy}. As illustrated in Figure \ref{fig:intro2}, it allows for flexible constraint on overestimated values of trajectories that receive low values from the ``knowledgeable'' source critic. Conversely, if a policy yields high values from the source critic, we prefer to retain the original estimation by the offline agent.
This approach is feasible because the source critic has been aligned with the target domain during world model learning.

We showcase the effectiveness of \model{} in offline visual control tasks across the Meta-World, RoboDesk, and DeepMind Control benchmarks. Our approach is shown to be readily extendable to scenarios with multiple source domains. It effectively addresses value overestimation by transferring knowledge from auxiliary domains, even in the presence of diverse physical dynamics, action spaces, reward scales, and visual appearances.
In summary, our work brings the following contributions:
\begin{itemize}[leftmargin=*]
\vspace{-5pt}
\item  We innovatively frame offline visual RL as a domain transfer problem. The fundamental idea is to harness cross-domain knowledge to tackle representation overfitting and value overestimation in offline visual control tasks.
\vspace{-2pt}
\item  We present \model{}, a method that follows the offline-online-offline paradigm, incorporating specific techniques of world model alignment and flexible value constraints.
\vspace{-5pt}
\end{itemize}

%% file: sections/02.problem_setup.tex
\section{Problem Setup}
\label{sec:problem_setup}

We consider offline visual reinforcement learning as a partially observable Markov decision process (POMDP) that aims to maximize the cumulative reward in a fixed target dataset $\mathcal{B}^{(T)}$.
We specifically focus on scenarios where auxiliary environments are accessible, enabling rich interactions and efficient online data collection.
The goal is to improve the offline performance of the target POMDP $\left\langle\mathcal{O}^{(T)}, \mathcal{A}^{(T)}, \mathcal{T}^{(T)}, \mathcal{R}^{(T)}, \gamma^{(T)} \right\rangle$ through knowledge transfer from the source POMDPs $\left\langle\mathcal{O}^{(S)}, \mathcal{A}^{(S)}, \mathcal{T}^{(S)}, \mathcal{R}^{(S)}, \gamma^{(S)} \right\rangle$. 
These notations respectively denote the space of visual observations, the space of actions, the state transition probabilities, the reward function, and the discount factor.

\begin{table*}[t]
\caption{RoboDesk (\textit{target domain}) vs. Meta-World (\textit{auxiliary source domain}).} 
\label{tab:meta_robo_cmp}
\setlength\tabcolsep{5pt}
\begin{center}
\footnotesize
\centering
\begin{tabular}{l|ccc}
\toprule
& Source: \textit{Meta-World} & Target: \textit{RoboDesk} &  Similarity / Difference \\
\midrule
Task& Window Close & Open Slide & Related manipulation tasks\\
Dynamics& Simulated Sawyer robot arm & Simulated Franka robot arm  & Different \\
Action space & Box(-1, 1, (4,), float64) & Box(-1, 1, (5,), float32) & Different\\
Reward scale& [0, 1]  & [0, 10] &  Different \\  
Observation & Right-view images  & Top-view images  & Different view points\\ 
\bottomrule
\end{tabular}
\end{center}
\vspace{-10pt}
\end{table*}

For example, in one of our experiments, we employ RoboDesk as the offline target domain and various tasks from Meta-World as the source domains.
As illustrated in Table \ref{tab:meta_robo_cmp}, these two environments present notable distinctions in physical dynamics, action spaces, reward definitions, and visual appearances as the observed images are from different camera views.
Our priority is to address domain discrepancies to enable cross-domain behavior learning.

%% file: sections/03.method.tex
\section{Method}

In this section, we present the technical details of \model{}, which consists of a pair of world models $\{\mathcal{M}_{\phi^{\prime}}, \mathcal{M}_\phi\}$, actor networks $\{\pi_{\psi^{\prime}}, \pi_{\psi}\}$, and critic networks $\{v_{\xi^{\prime}}, v_\xi\}$, where $\{\phi, \psi, \xi\}$ and $\{\phi^{\prime}, \psi^{\prime}, \xi^{\prime}\}$ are respectively target and source domain parameters.
As potential cross-domain discrepancies may exist in all elements of $\{\mathcal{O}, \mathcal{A}, \mathcal{T}, \mathcal{R}\}$, the entire training process is organized into three iterative stages, following an \textit{offline-online-offline} transfer learning framework:
\begin{enumerate}
\renewcommand{\labelenumi}{\Alph{enumi})}
    \vspace{-5pt}
    \item \textit{Offline-to-online state alignment}: Train the offline world model $\mathcal{M}_\phi$ by aligning its state space with that of the source world model $\mathcal{M}_{\phi^{\prime}}$.
    \vspace{-2pt}
    \item \textit{Online-to-offline reward alignment}: Train $\mathcal{M}_{\phi^{\prime}}$ and $\{\pi_{\psi^{\prime}},v_{\xi^{\prime}}\}$ in the online environment by incorporating the target reward information. 
    \vspace{-2pt}
    \item \textit{Online-to-offline value constraint}: Train the target offline-domain agent $\{\pi_\psi,v_\xi\}$ with value constraints provided by the source critic $v_{\xi^{\prime}}$.
\end{enumerate}

\subsection{Offline-to-Online State Alignment}
\label{sec:target_model}

\paragraph{Source model pretraining.} 
We start with a source domain warm-up phase employing a model-based actor-critic method known as DreamerV2~\cite{hafner2021mastering}. 
To facilitate cross-domain knowledge transfer, we additionally introduce a state alignment module, which is denoted as $g(\cdot)$ and implemented using the softmax operation.
The world model $\mathcal{M}_{\phi^{\prime}}$ consists of the following components:
\begin{equation}
\label{eq:wm}
\small
\begin{alignedat}{8}
&\text{Recurrent transition: } && h_t^{(S)} = f_{\phi^\prime}(h_{t-1}^{(S)}, z_{t-1}^{(S)}, a_{t-1}^{(S)}) && \
\; 
\text{Image encoding: } && {e}_t^{(S)} = e_{\phi^{\prime}}(o_t^{(S)})\\
&\text{Posterior state: } && z_t^{(S)} \sim q_{\phi^{\prime}}(h_t^{(S)}, {e}_t^{(S)}) && \
\; 
\text{Prior state: } && \hat{z}_t^{(S)} \sim p_{\phi^{\prime}}(h_t^{(S)}) \\ 
&\text{Reconstruction: } && \hat{o}_t^{(S)} \sim p_{\phi^\prime}(h_t^{(S)},z_t^{(S)}) && \
\; 
\text{Reward prediction: } && \hat{r}_t^{(S)} \sim r_{\phi^{\prime}}(h_t^{(S)},z_t^{(S)}) \\
&\text{Discount factor: } &&\hat{\gamma}^{(S)}_t \sim p_{\phi^\prime}(h_t^{(S)},z_t^{(S)}) && \
\; 
\text{State alignment target: \ } && s_t^{(S)} = g({e}_t^{(S)}), \\
\end{alignedat}
\end{equation}
where $\phi^\prime$ represents the combined parameters of the world model.
We train $\mathcal{M}_{\phi^{\prime}}$ on the dynamically expanded source domain experience replay buffer $\mathcal{B}^{(S)}$ by minimizing 
\begin{equation}
\small
\begin{aligned}
\label{eq:source_world_model_loss}
\mathcal{L}(\phi^{\prime}) = \ & \mathbb{E}_{q_{\phi^{\prime}}}
\Big[
\sum_{t=1}^N \underbrace{-\ln p_{\phi^{\prime}}(o_t^{(S)} \mid h_t^{(S)}, z_t^{(S)})}_{\text {image reconstruction}} \underbrace{-\ln r_{\phi^{\prime}}(r_t^{(S)} \mid h_t^{(S)}, z_t^{(S)})}_{\text {reward prediction}} \underbrace{-\ln p_{\phi^{\prime}}(\gamma_t^{(S)} \mid h_t^{(S)}, z_t^{(S)})}_{\text {discount prediction }} \\ &\underbrace{+ \ \mathrm{KL}\left[q_{\phi^{\prime}}(z_t^{(S)} \mid h_t^{(S)}, o_t^{(S)}) \ \| \ p_{\phi^{\prime}}(\hat{z}_t^{(S)} \mid h_t^{(S)})\right]}_{\text{KL divergence}}\Big].
\end{aligned}    
\end{equation}
We train the source actor $\pi_{\psi^\prime}(\hat{z}_t)$ and critic $v_{\xi^\prime}(\hat{z}_t)$ with the respective objectives of maximizing and estimating the expected future rewards $\mathbb{E}_{p_{\phi^\prime},p_{\psi^\prime}}[\sum_{\tau \geq t}\hat{\gamma}_{\tau-t}\hat{r}_\tau]$ generated by $\mathcal{M}_{\phi^{\prime}}$. Please refer to \underline{Appendix \ref{sec:bl}} for more details.
We deploy $\pi_{\psi^\prime}$ to interact with the auxiliary environment and collect new data for further world model training. 

\vspace{-5pt}
\paragraph{State alignment.} 
A straightforward transfer learning solution is to train the target agent in the offline dataset upon the checkpoints of the source agent. However, it may suffer from a potential mismatch issue due to the discrepancy in tasks, visual observations, physical dynamics, and action spaces across various domains.
This becomes more severe when the online data is collected from environments that differ from the offline dataset (\textit{e.g.}, Meta-World $\rightarrow$ RoboDesk). 
We tackle this issue by separating the parameters of the source and the target agents while explicitly aligning their latent state spaces. 
Concretely, the target world model $\mathcal{M}_\phi$ has an identical network architecture to the source model $\mathcal{M}_{\phi^{\prime}}$.
We feed the same target domain observations sampled from $\mathcal{B}^{(T)}$ into these models and close the distance of $e_{\phi^{\prime}}(o_t^{(T)})$ and $e_{\phi}(o_t^{(T)})$.
%
We optimize $\mathcal{M}_\phi$ by minimizing
\vspace{-5pt}
\begin{equation}
\small
\begin{aligned}
\label{eq:target_world_model_loss}
\mathcal{L}(\phi) & = \mathbb{E}_{q_\phi}
\Big[
\sum_{t=1}^N \underbrace{-\ln p_\phi(o_t^{(T)} \mid h_t^{(T)}, z_t^{(T)})}_{\text {image reconstruction }} \underbrace{-\ln r_\phi(r_t^{(T)} \mid h_t^{(T)}, z_t^{(T)})}_{\text {reward prediction}} 
\underbrace{-\ln p_\phi(\gamma_t^{(T)} \mid h_t^{(T)}, z_t^{(T)})}_{\text {discount prediction}} \\
&\underbrace{+ \ \beta_1 \mathrm{KL}\left[q_\phi(z_t^{(T)} \mid h_t^{(T)}, o_t^{(T)}) \| \ p_\phi(\hat{z}_t^{(T)} \mid h_t^{(T)})\right]}_{\text{KL divergence}}
\underbrace{+ \ \beta_2 \mathrm{KL}\left[\texttt{sg}(g(e_{\phi^{\prime}}(o_t^{(T)}))) \ \| \ g(e_{\phi}(o_t^{(T)}))\right]}_{\text{domain alignment loss}}\Big],
\end{aligned}
\end{equation}
where \texttt{sg}($\cdot$) indicates gradient stopping and we use the encoding from the source model as the state alignment target.
As the source world model can actively interact with the online environment and gather rich information, it keeps the target world model from overfitting the offline data.
The importance of this loss term is governed by $\beta_2$. We examine its sensitivity in the experiments.

\begin{algorithm}[t]
  \caption{The training scheme of \model{}.}
  \label{algo:overall}
  \begin{algorithmic}[1]
  \small
  \State \textbf{Require: }{Offline dataset $\mathcal{B}^{(T)}$.} 
  \State \textbf{Initialize:} Parameters of the source model $\{\phi^{\prime}, \psi^{\prime}, \xi^{\prime}\}$ and the target model $\{\phi, \psi, \xi\}$. 
  \State Pretrain the source agent and collect a replay buffer $\mathcal{B}^{(S)}$.  
 \While{not converged}
    \For{each step in $\{1:K_1\}$} \Comment{In the offline domain}
        \State Sample $\{(o_{t}^{(T)}, a_{t}^{(T)}, r_{t}^{(T)})\}_{t=1}^{N} \sim \mathcal{B}^{(T)}$. 
        \State Train the target world model $\mathcal{M}_\phi$ using \eqref{eq:target_world_model_loss}. \Comment{Offline-to-online state alignment}
        \State Generate $\{(z_i^{(T)}, a_i^{(T)})\}_{i=t}^{t+H}$ using $\pi_\psi$ and $\mathcal{M}_\phi$. \Comment{Behavior learning with constraint} 
        \State Train the critic $v_\xi$ using \eqref{eq:critic_loss} over $\{(z_i^{(T)}, a_i^{(T)})\}_{i=t}^{t+H}$. 
        \State Train the actor $\pi_\psi$ using \eqref{eq:actor_loss} over $\{(z_i^{(T)}, a_i^{(T)})\}_{i=t}^{t+H}$. 
     \EndFor
     
    \For{each step in $\{1:K_2\}$} \Comment{In the online domain}
        \State Sample $\{(o_{t}^{(S)}, a_{t}^{(S)}, r_{t}^{(S)})\}_{t=1}^{N} \sim \mathcal{B}^{(S)}$. 
        \State Sample $\{(o_{t}^{(T)}, a_{t}^{(T)}, r_{t}^{(T)})\}_{t=1}^{N} \sim \mathcal{B}^{(T)}$. \Comment{Online-to-offline reward alignment}
        \State Relabel the source rewards $\{\tilde{r}_t^{(S)}\}_{t=1}^{N}$ using \eqref{eq:corrected_reward}.
        \State Train $\mathcal{M}_{\phi^{\prime}}$ using \eqref{eq:source_world_model_loss} combined with \eqref{eq:mleloss}.
        \State Generate $\{(z_i^{(S)}, a_i^{(S)})\}_{i=t}^{t+H}$ using $\pi_{\psi^{\prime}}$ and $\mathcal{M}_{\phi^{\prime}}$. \Comment{Source domain behavior learning}
        \State Train $\pi_{\psi^{\prime}}$ and $v_{\xi^{\prime}}$ over the imagined $\{(z_i^{(S)}, a_i^{(S)})\}_{i=t}^{t+H}$.
        \State Use $\pi_{\psi^{\prime}}$ to collect new source data and append $\mathcal{B}^{(S)}$. 
    \EndFor
    \EndWhile
\end{algorithmic}
\end{algorithm}

\subsection{Online-to-Offline Reward Alignment}
\label{sec:source_model}

To enable the source agent to value the target policy, it is essential to provide it with prior knowledge of the offline task.
To achieve this, we train the source reward predictor $r_{\phi^{\prime}}(\cdot)$ using mixed data from both of the replay buffers $\mathcal{B}^{(S)}$ and $\mathcal{B}^{(T)}$.
Through the behavior learning on source domain imaginations, the target-informed reward predictor enables the source RL agent to assess the imagined states produced by the target model and provide a flexible constraint to target value estimation (as we will discuss in Section \ref{sec:target_behavior}).

Specifically, we first sample a target domain data trajectory $\{(o_{t}^{(T)}, a_{t}^{(T)}, r_{t}^{(T)})\}_{t=1}^{T}$ from $\mathcal{B}^{(T)}$ (\textbf{Line 14} in Alg. \ref{algo:overall}). 
We then use the source world model parametrized by $\phi^{\prime}$ to extract corresponding latent states and relabel the \textit{target-informed source reward} (\textbf{Line 15} in Alg. \ref{algo:overall}): 
\begin{equation}
\small
\begin{aligned}
\label{eq:corrected_reward}
    & \tilde{h}_t = f_{\phi^\prime}(\tilde{h}_{t-1}, \tilde{z}_{t-1}, a_{t-1}^{(T)}) &\quad
    & \tilde{e}_t = e_{\phi^{\prime}}(o_t^{(T)}) \\
    & \tilde{z}_t \sim q_{\phi^{\prime}}(\tilde{h}_t, \tilde{e}_t) &\quad
    & \tilde{r}_{t}^{(S)} = (1-k) \cdot r_{\phi^{\prime}}(\tilde{h}_t, \tilde{z}_t) + k \cdot r^{(T)}_t,
\end{aligned}
\end{equation}
where $k$ is the target-informed reward factor, which acts as a balance between the true target reward $r^{(T)}_t$ and the output of the source reward predictor ${r}_{\phi^{\prime}}(\cdot)$ provided with target states.
It is crucial to emphasize that using the target data as inputs to compute ${r}_{\phi^{\prime}}(\cdot)$ is feasible due to the alignment of the target state space with the source state space.

We jointly use the relabeled reward $\tilde{r}_{t}^{(S)}$ and the original source domain reward $r_{t}^{(S)}$ sampled from $\mathcal{B}^{(S)}$ to train the source reward predictor. This training is achieved by minimizing a maximum likelihood estimation (MLE) loss:
\begin{equation}
\small
\begin{aligned}
    \label{eq:mleloss}
    \mathcal{L}_r(\phi^{\prime}) 
    = \ \eta \cdot \mathbb{E}_{\mathcal{B}^{(S)}}
    \Big[\sum_{t=1}^N -\ln r_{\phi^{\prime}}(r_t^{(S)} | h_t^{(S)}, z_t^{(S)} )\Big] +(1-\eta) \mathbb{E}_{\mathcal{B}^{(T)}}
    \Big[\sum_{t=1}^N -\ln r_{\phi^{\prime}}(\tilde{r}_t^{(S)} | h_t^{(T)}, z_t^{(T)})\Big],
\end{aligned}
\end{equation}
where the second term measures the negative log-likelihood of observing the relabelled source reward $\tilde{r}_t^{(S)}$. 
$\eta$ represents a hyperparameter that gradually decreases from $1$ to $0.1$ throughout this training stage. 
Intuitively, $\eta$ controls the progressive adaptation of the well-trained source reward predictor to the target domain with limited target reward supervision.
We integrate \eqref{eq:mleloss} into \eqref{eq:source_world_model_loss} to train the entire world model $\mathcal{M}_{\phi^{\prime}}$ for the source domain agent (\textbf{Line 16} in Alg. \ref{algo:overall}) and subsequently perform behavior learning to enable the source critic to assess the target policy (\textbf{Lines 17-19} in Alg. \ref{algo:overall}).

\subsection{Min-Max Value Constraint}
\label{sec:target_behavior}

In the behavior learning phase of the target agent (\textbf{Lines 8-10} of Alg. \ref{algo:overall}), we mitigate value overestimation in the offline dataset by introducing a min-max regularization term to the objective function of the target critic model $v_\xi$.  
Initially, we use the auxiliary source critic $v_{\xi^{\prime}}$ to estimate the value function of the imagined target states. 
Following that, we train $v_\xi$ by additionally \emph{minimizing the maximum value} among the estimates provided by source and target critics: 
\begin{equation}
\label{eq:critic_loss}
\small
\begin{aligned}
    \mathcal{L}(\xi) =  \ \mathbb{E}_{p_{\phi},p_{\psi}} \Big[ \sum^{H-1}_{t=1}\underbrace{\frac{1}{2}\left(v_{\xi}(\hat{z}_t^{(T)})-\texttt{sg}\big(V_t^{(T)}\big)\right)^2}_{\text{value regression}} 
     + \ \underbrace{\alpha \max\left(v_{\xi}(\hat{z}_t^{(T)}), \ \texttt{sg}\big(v_{\xi^{\prime}}(\hat{z}_t^{(T)})\big)\right)}_{\text{value constraint}}\Big],
\end{aligned}
\end{equation}
where $V_t^{(T)}$ incorporates a weighted average of reward information over an $n$-step future horizon.
The first term in the provided loss function fits cumulative value estimates (whose specific formulation can be located in \underline{Appendix \ref{sec:bl}}), while the second term regularizes the overestimated values for out-of-distribution data in a mildly conservative way. 
The hyperparameter $\alpha$ represents the importance of the value constraint.
The \texttt{sg}($\cdot$) operator indicates that we stop the gradient to keep the source critic from being influenced by the regularization term.

This approach provides flexibly conservative value estimations, finding a balance between mitigating overestimation and avoiding excessive conservatism in the value function.
When the target critic overestimates the value function, the source critic is less vulnerable to the value overestimation problem as it is trained with rich interaction data. Thus, it is possible to observe $v_\xi(\hat{z}_t^{(T)}) > v_{\xi^{\prime}}(\hat{z}_t^{(T)})$, and our approach is designed to decrease the output of $v_\xi$ to the output of $v_{\xi^{\prime}}$. This prevents the target critic from overestimating the true value. 
Conversely, when the source critic produces greater values in $v_{\xi^{\prime}}(\hat{z}_t^{(T)})$, the min-max regularization term does not contribute to the training of the target critic $v_{\xi}$. This encourages the exploration of potentially advantageous states within the imaginations of the target world model.
In line with DreamerV2 \cite{hafner2021mastering},
we train the target actor $\pi_\psi$ by maximizing a REINFORCE objective function with entropy regularization, allowing the gradients to backpropagate directly through the learned dynamics:
\begin{equation}
\small
\label{eq:actor_loss}
\begin{aligned}
    \mathcal{L}(\psi) =  \mathbb{E}_{p_{\phi}, p_{\psi}}
    \sum_{t=1}^{H-1}(
    \underbrace{\beta \mathrm{H}[a_t^{(T)} \mid \hat{z}_t^{(T)}]}_{\text{entropy regularization}}
    + \underbrace{\rho V_t^{(T)}}_{\text{dynamics backprop}} 
    + \underbrace{(1-\rho) \ln \pi_\psi(\hat{a}_t^{(T)} \mid \hat{z}_t^{(T)}) \texttt{sg}(V_t^{(T)}-v_{\xi}(\hat{z}_t^{(T)})}_{\text{REINFORCE}}).
\end{aligned}
\end{equation}
As previously mentioned, $V_t^{(T)}$ involves a weighted average of reward information over an $n$-step future horizon, with detailed formulation provided in \underline{Appendix \ref{sec:bl}}.


Furthermore, it is crucial to note that \model{} can readily be extended to scenarios with multiple source domains by adaptively selecting a useful task as the auxiliary domain. This extension is easily achieved by measuring the distance of the latent states between the target domain and each source domain. 
For technical details of the adaptive source domain selection, please refer to \underline{Appendix \ref{sec:multi_source}}.

%% file: sections/04.experiments.tex
\section{Experiments}
\label{sec:exp}

\begin{table*}[t]
\vspace{-5pt}
\caption{Mean episode returns and standard deviations of $10$ episodes over $3$ seeds on Meta-World. 
}
\label{tab:metaworld_result}
\setlength\tabcolsep{2.8pt}
\footnotesize
\begin{center}
\begin{tabular}{l|ccccccc}
\toprule
Model & BP$\rightarrow \text{DC}^{*}$ 
& DC $\rightarrow$ BP
& BT$\rightarrow$ WC
& BP$\rightarrow$ HP
& WC$ \rightarrow$ DC
& HP$\rightarrow$ BT
& Avg.
\\
\midrule
Offline DV2 & 2143$\pm$579 & 3142$\pm$533 & 3921$\pm$752 & 278$\pm$128 & 3899$\pm$679 & 3002$\pm$346 & 2730 \\
DrQ + BC& 567$\pm$19 & 587$\pm$68 & 623$\pm$85 & 1203$\pm$234 & 134$\pm$64 & 642$\pm$99 & 626\\
CQL& 1984$\pm$13 & 867$\pm$330 & 683$\pm$268 & 988$\pm$39 & 577$\pm$121 & 462$\pm$67 & 927\\
CURL& 1972$\pm$11 & 51$\pm$17 & 281$\pm$73 & 986$\pm$47  & 366$\pm$52 & 189$\pm$10& 641\\
LOMPO  &2883$\pm$183 & 446$\pm$458  & 2983$\pm$569 & 2230$\pm$223 & 2756$\pm$331 & 1961$\pm$287 & 1712\\
\cmidrule{1-8}
DV2 Finetune & 3500$\pm$414 & 2456$\pm$661 & 3467$\pm$1031 & 3702$\pm$451 & 4273$\pm$1327 & 3499$\pm$713 & 3781\\
DV2 Finetune + EWC & 1566$\pm$723 & 167$\pm$86 & 978$\pm$772 & 528$\pm$334  & 2048$\pm$1034 & 224$\pm$147 & 918\\
LOMPO Finetune & 259$\pm$191 & 95$\pm$53 &142$\pm$70 &332$\pm$452 & 3698$\pm$1615 & 224$\pm$88 & 792\\
\cmidrule{1-8}
\model{} (Best-Source) &\bfseries 3967$\pm$312 & \bfseries 3623$\pm$543  & \bfseries 4521$\pm$367 &\bfseries 4570$\pm$677 & \bfseries4845$\pm$14 & \bfseries 3889$\pm$159  & \bfseries 4241\\
\model{} (Multi-Source) & \underline{3864$\pm$352} & \underline{3573$\pm$541} & \underline{4507$\pm$59} & \underline{4460$\pm$783} & \underline{4678$\pm$137} & \underline{3626$\pm$275} & \underline{4094}\\
\bottomrule
\end{tabular}
\end{center}
\vspace{-10pt}
\end{table*}

\subsection{Experimental Setups}
\label{sec:exp_set}

\paragraph{Datasets.} 
We evaluate \model{} across three visual control environments, \textit{i.e.}, Meta-World~\citep{yu2019meta}, RoboDesk~\citep{kannan2021robodesk}, and DeepMind Control Suite (DMC)~\citep{tassa2018deepmind}, including both cross-task and cross-environment setups (Meta-World $\rightarrow$ RoboDesk). 
Inspired by D4RL~\citep{fu2020d4rl}, we build offline datasets of \textit{medium-replay} quality using DreamerV2~\cite{hafner2021mastering}.
The datasets comprise all the samples in the replay buffer collected during the training process until the policy attains medium-level performance, defined as achieving $1/3$ of the maximum score that the DreamerV2 agent can achieve.
Please refer to \underline{Appendix \ref{sec:dmc_expert_res}} for further results of \model{} trained with \textit{medium-expert} offline data.

\vspace{-5pt}
\paragraph{Compared methods.}
We compare \model{} with both model-based and model-free RL approaches, including \textit{Offline DV2} \citep{lu2023challenges}, \textit{DrQ+BC} \citep{lu2023challenges}, \textit{CQL} \citep{lu2023challenges},
\textit{CURL} \citep{laskin2020curl}, and \textit{LOMPO} \citep{rafailov2021offline}. 
In addition, we introduce the \textit{DV2 Finetune} method, which involves taking a DreamerV2~\citep{hafner2021mastering} model pretrained in the online source domain and subsequently finetuning it in the offline target dataset. 
Furthermore, \textit{DV2 Finetune} can be integrated with the continual learning method, Elastic Weight Consolidation~(EWC)~\citep{kirkpatrick2017overcoming}, to regularize the model for preserving source domain knowledge, \textit{i.e.}, \textit{Finetune+EWC}. Please refer to \underline{Appendix \ref{sec:compared_methods}} for more details.

\subsection{Cross-Task Experiments on Meta-World}

Meta-World is an open-source simulated benchmark designed for solving a wide range of robot manipulation tasks. We select $6$ tasks as either the offline dataset or potential candidates for the online auxiliary domain. These tasks include: \textit{Door Close}~($\textbf{DC}^{*}$), \textit{Button Press}~(\textbf{BP}),  \textit{Window Close}~(\textbf{WC}), \textit{Handle Press}~(\textbf{HP}), \textit{Drawer Close}~(\textbf{DC}), \textit{Button Topdown}~(\textbf{BT}).

\vspace{-8pt}
\paragraph{Main results.}
As shown in \tabref{tab:metaworld_result}, we compare the results of \model{} with other models on Meta-World. \model{} achieves the best performance in all $6$ tasks. Notably, it outperforms \textit{Offline DV2} \citep{lu2023challenges}, a method also built upon DreamerV2 and specifically designed for offline visual RL.
For the online-to-offline finetuning models, \textit{DV2 Finetune} achieves the second-best results by leveraging transferred knowledge from the auxiliary source domain. 
However, we observe that its performance experiences a notable decline in scenarios (\textit{e.g.}, Meta-World $\rightarrow$ RoboDesk) involving significant data distribution shifts between the source and the target domains in visual observation, physical dynamics, reward definition, or even the action space of the robots.
Another important baseline model is \textit{DV2 Finetune+EWC}, which focuses on mitigating the catastrophic forgetting of the knowledge obtained in source domain pretraining. Nevertheless, without additional model designs for domain adaptation, retaining source domain knowledge may eventually lead to a decrease in performance in the target domain.
The LOMPO model suffers from the \textit{negative transfer} effect when incorporating a source pretraining stage. It achieves an average return of $1{,}712$ when it is trained from scratch in the offline domain while achieving an average return of $792$ for online-to-offline finetuning. It implies that a na\"ive transfer learning method may degenerate the target performance due to unexpected bias.

\begin{figure*}[t]
    \centering
\includegraphics[width=0.92\textwidth]{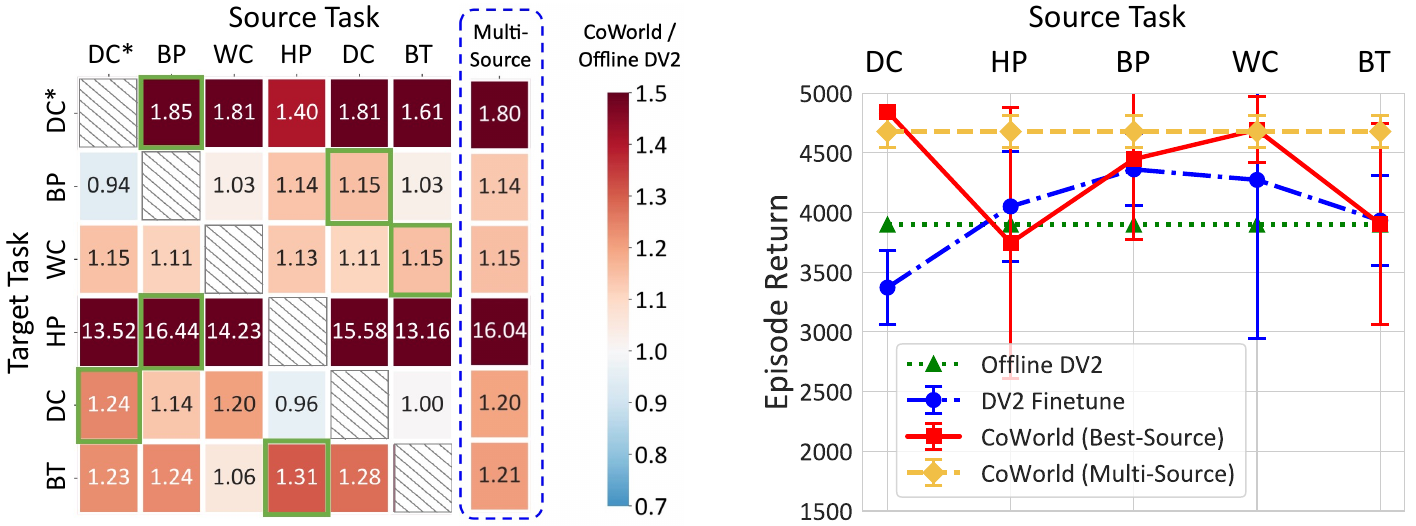}
  \vspace{-5pt}
    \caption{\textbf{Left:} The value in each grid indicates the ratio of returns achieved by \model{} compared to \textit{Offline DV2}. 
    Highlighted grids represent the top-performing source domain. 
    \textbf{Right:} Returns on \textit{Drawer Close} (DC*) with different source domains, where the multi-source \model{} (yellow line) is shown to automatically discover (\textit{i.e.}, \textit{Door Close}) as the source domain and achieve comparable results with the top-performing single-source \model{} (red line).}
    \label{fig:metaworld_fintune_result}
\end{figure*}

\begin{figure*}[t]
    \centering
\includegraphics[width=\textwidth]{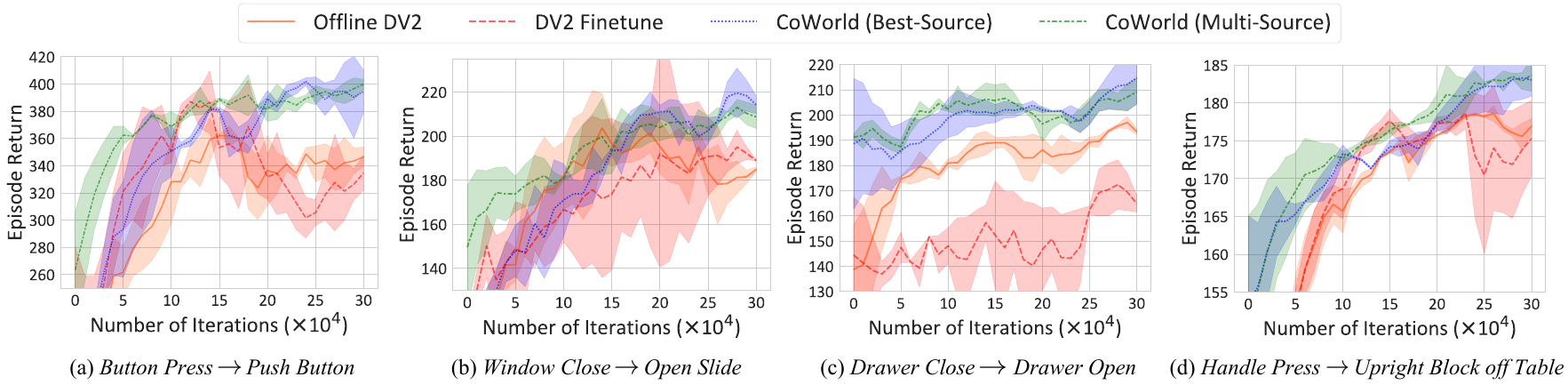}
  \vspace{-15pt}
    \caption{Quantitative results in domain transfer scenarios of Meta-World $\rightarrow$ RoboDesk.}
    \label{fig:robodesk_result}
    \vspace{-10pt}
\end{figure*}

\vspace{-5pt}
\paragraph{Results with a random source domain.}
Given that we present the \textit{best-source} results in \tabref{tab:metaworld_result}, where we manually select one source task from Meta-World, one may cast doubt on the influence of domain discrepancies between the auxiliary environment and the target offline dataset.
In \figref{fig:metaworld_fintune_result} (Left), the transfer matrix of \model{} among the $6$ tasks of Meta-World is presented, where values greater than $1$ indicate positive domain transfer effects. Notably, there are challenging cases with weakly related source and target tasks. In the majority of cases ($26$ out of $30$), \model{} outperforms \textit{Offline DV2}, as illustrated in the heatmap.

\vspace{-5pt}
\paragraph{Results with multiple source domains.}
It is crucial to note that \model{} can be easily extended to scenarios with multiple source domains by adaptively selecting a useful task as the auxiliary domain. 
From \tabref{tab:metaworld_result}, we can see that the multi-source CoWorld achieves comparable results to the models trained with manually designated online simulators. 
In \figref{fig:metaworld_fintune_result} (Left), multi-source \model{} achieves positive improvements over \textit{Offline DV2} in all cases, approaching the best results of models using each source task as the auxiliary domain. 
In \figref{fig:metaworld_fintune_result} (Right), it also consistently outperforms the \textit{DV2 Finetune} baseline model.
These results demonstrate our approach's ability to execute without strict assumptions about domain similarity and its ability to automatically identify a useful online simulator from a set of both related and less related source domains.

\subsection{Cross-Environments: Meta-World to RoboDesk}
\label{sec:robodesk_analysis}

To explore cross-environment transfer with more significant domain gaps, we employ four tasks from RoboDesk to construct individual offline datasets, \ie \textit{Push Button}, \textit{Open Slide}, \textit{Drawer Open}, \textit{Upright Block off Table}.
These tasks require handling randomly positioned objects with image inputs.
Table \ref{tab:meta_robo_cmp} presents the differences between the two environments in physical dynamics, action space, reward definitions, and visual appearances.

\figref{fig:robodesk_result} presents quantitative comparisons, where \model{} outperforms \textit{Offline DV2} and \textit{DV2 Finetune} by large margins.
For the \textit{best-source} experiments, we manually select one source domain from Meta-World. 
For the \textit{multi-source} experiments, we jointly use all Meta-World tasks as the source domains.
In contrast to prior findings, directly finetuning the source world model in this cross-environment setup, where there are more pronounced domain discrepancies, does not result in significant improvements in the final performance.
In comparison, \model{} more successfully addresses these challenges by leveraging domain-specific world models and RL agents, and explicitly aligning the state and reward spaces across domains.
We also showcase the performance of multi-source \model{}, which achieves comparable results to the \textit{best-source} model that exclusively uses our designated source domain.

\subsection{Cross-Dynamics Experiments on DMC}

DMC is a widely explored benchmark for continuous control. We use the \textit{Walker} and \textit{Cheetah} as the base agents and make modifications to the environment to create a set of $8$ distinct tasks, \textit{i.e.}, \textit{Walker Walk}~(\textbf{WW}), \textit{Walker Downhill}~(\textbf{WD}), \textit{Walker Uphill}~(\textbf{WU}), \textit{Walker Nofoot}~(\textbf{WN}), \textit{Cheetah Run}~(\textbf{CR}), \textit{Cheetah Downhill}~(\textbf{CD}), \textit{Cheetah Uphill}~(\textbf{CU}), \textit{Cheetah Nopaw}~(\textbf{CN}).
Particularly, \textit{Walker Nofoot} is a task in which we cannot control the right foot of the \textit{Walker} agent. \textit{Cheetah Nopaw} is a task in which we cannot control the front paw of the \textit{Cheetah} agent.

\begin{figure*}[!t]
    \centering
    \subfigure[Meta-World $\rightarrow$  RoboDesk (\textit{Push Green Button})
    ]{\includegraphics[height=0.29\textwidth]{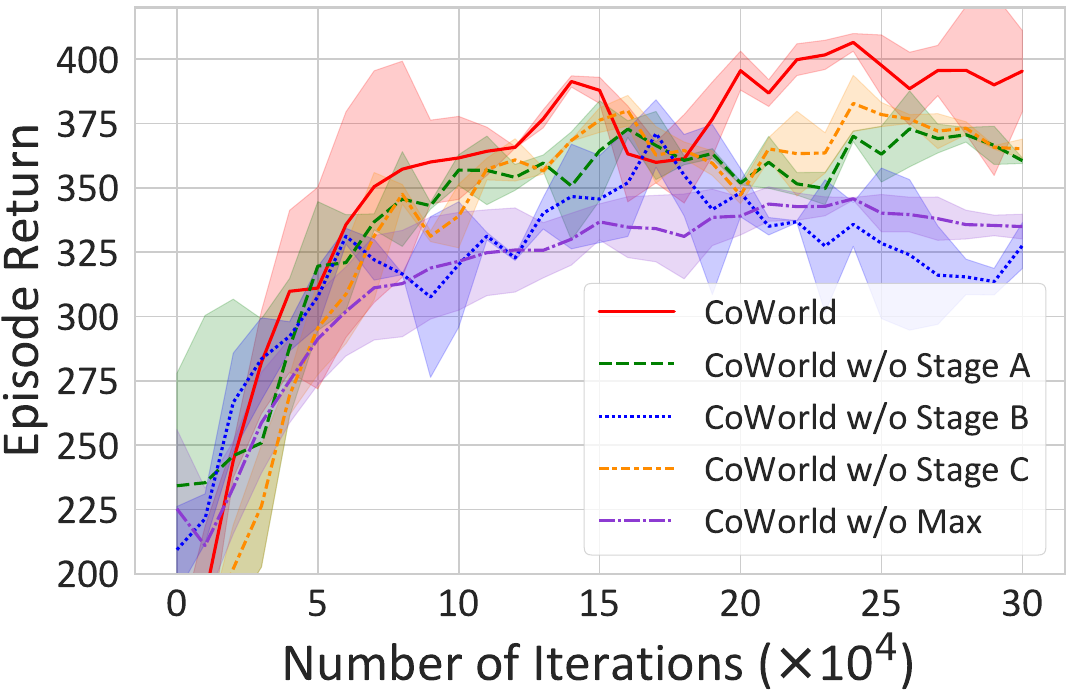}
    \label{fig:ablation}
    }
    \subfigure[Y-axis $=$ Estimated value $-$ True value]{\includegraphics[height=0.28\textwidth]{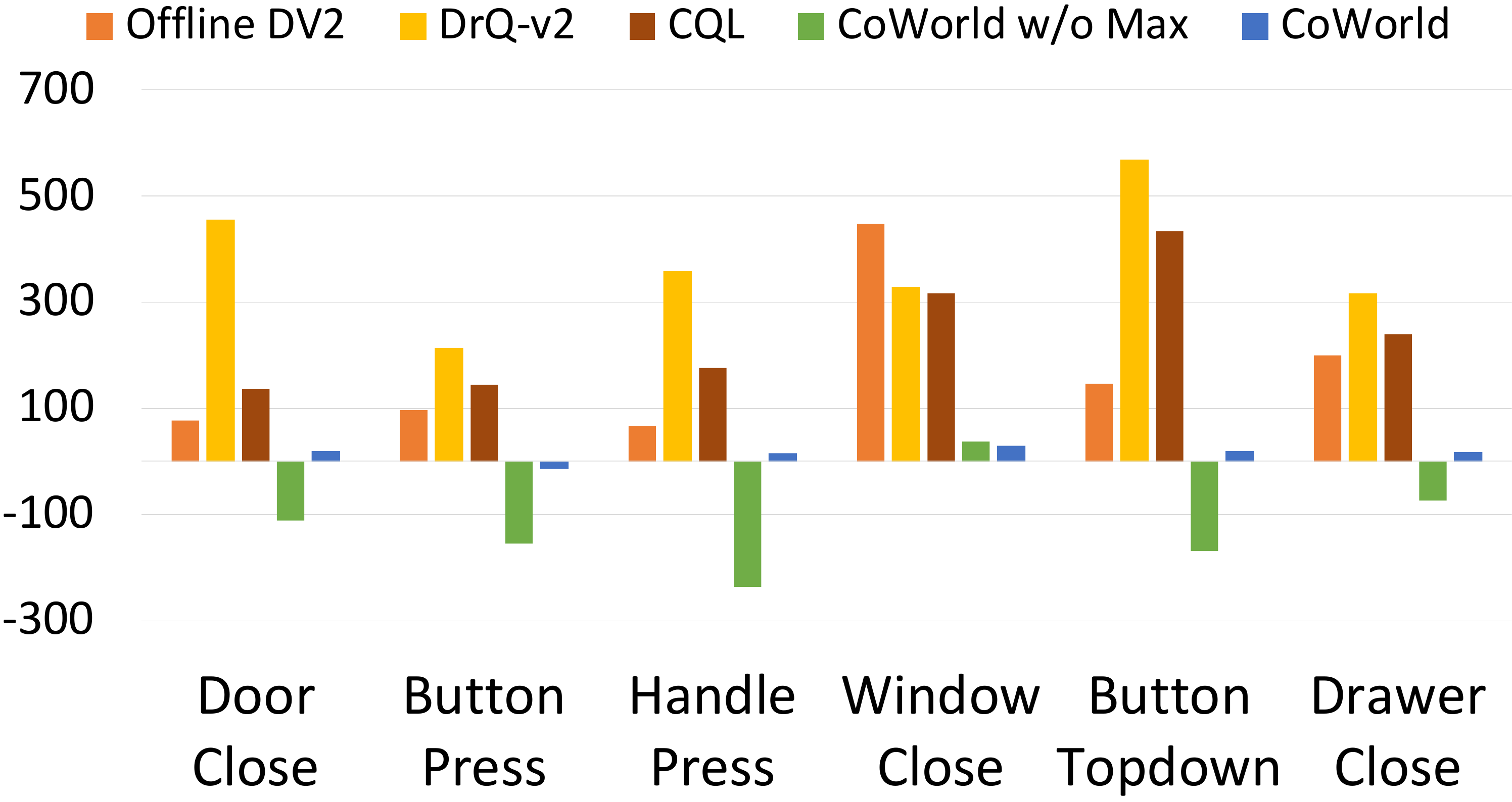}
    \label{fig:Q-value}
    } 
    \vspace{-8pt}
    \caption{
    \textbf{(a)} Ablation studies on \textcolor{MyDarkGreen}{state alignment}, \textcolor{MyDarkBlue}{reward alignment}, and \textcolor{orange}{min-max value constraint}.
    \textbf{(b)} The disparities between the estimated value by various models and the true value. 
    Please see the text in Section \ref{rec:ablation} for the implementation of \textcolor{MyPurple}{\model{} w/o Max}.
    }
    \label{fig:some_results}
\end{figure*}

\begin{table*}[t]
\caption{Mean rewards and standard deviations of $10$ episodes in offline DMC over $3$ seeds. 
} 
\label{tab:dmc_result}
\setlength\tabcolsep{2.8pt}
\begin{center}
\footnotesize
\begin{tabular}{l|ccccccc}
\toprule
Model& WW $\rightarrow$ WD & WW $\rightarrow$ WU &  WW $\rightarrow$ WN  & CR $\rightarrow$ CD & CR $\rightarrow$ CU & CR $\rightarrow$ CN & Avg.
\\
\midrule
Offline DV2& 435$\pm$22 & 139$\pm$4 &  214$\pm$4  & 243$\pm$7 & 3$\pm$1 & 51$\pm$4 &181 \\
DrQ+BC& 291$\pm$ 10& 299$\pm$15  & 318$\pm$40  & 663$\pm$15 & 202$\pm$12 & 132$\pm$33 &355\\
CQL& $46\pm$19 &  64$\pm$32 & 29$\pm$2  & 2$\pm$1 & 52 $\pm$57& 111$\pm$157 &51\\
CURL& 43$\pm$5 &  21$\pm$3 & 23$\pm$3  &  26$\pm$7& 4$\pm$2 & 11$\pm$4 &21\\
LOMPO  & \underline{462$\pm$87} &  260$\pm$21 & \bfseries460$\pm$9  & 395$\pm$52& 46$\pm$19 & 120$\pm$4 &291 \\
\cmidrule{1-8}
DV2 Finetune & 379$\pm$23 &  \underline{354$\pm$29} & 407$\pm$37  &  \underline{702$\pm$41}& \underline{208$\pm$22} & \underline{454$\pm$82} & \underline{417}\\
LOMPO Finetune &209$\pm$21 &141$\pm$27 &212$\pm$9 &142$\pm$29 &17$\pm$11& 105$\pm$12 &137
\\
\model{} & \bfseries 629$\pm$9 &  \bfseries 407$\pm$141 &  \underline{426$\pm$32}  & \bfseries 745$\pm$28&\bfseries 225$\pm$20 & \bfseries 493$\pm$10 &\bfseries 488 \\ 
\bottomrule
\end{tabular}
\end{center}
\vspace{-10pt}
\end{table*}

We apply the proposed multi-source domain selection method to build the domain transfer settings shown in Table \ref{tab:dmc_result}. It is worth noting that \model{} outperforms the other compared models in $5$ out of $6$ DMC offline datasets, and achieves the second-best performance in the remaining task. On average, it outperforms \textit{Offline DV2} by $169.6\%$ and outperforms \textit{DrQ+BC} by $37.5\%$. Corresponding qualitative comparisons can be found in \underline{Appendix \ref{sec:vis_policy_evaluation}}.

\subsection{Further Analyses}
\label{rec:ablation}

\paragraph{Ablation studies.}
We conduct a series of ablation studies to validate the effectiveness of state space alignment ({Stage A}), reward alignment ({Stage B}), and min-max value constraint ({Stage C}).
We show corresponding results on the offline \textit{Push Green Button} dataset from RoboDesk in \figref{fig:ablation}. The performance experiences a significant decline when we abandon each training stage in \model{}. 

\vspace{-5pt}
\paragraph{Can \model{} address value overestimation?}
We evaluate the values estimated by the critic network of \model{} on the offline Meta-World datasets when the training process is finished.
In \figref{fig:Q-value}, we compute the cumulative value predictions throughout $500$ steps. 
The \textit{true value} is determined by calculating the discounted sum of the actual rewards obtained by the actor in the same $500$-steps period. 
We observe that existing approaches, including \textit{Offline DV2} and \textit{CQL}, often overestimate the value functions in the offline setup. 
The baseline model ``\textit{\model{} w/o Max}'' is a variant of \model{} that incorporates a brute-force constraint on the critic loss. It reformulates \eqref{eq:critic_loss} as $\sum^{H-1}_{t=1}\frac{1}{2}(v_{\xi}(\hat{z}_t)-\texttt{sg}(V_t))^2 + \alpha v_{\xi}(\hat{z}_t)$. As observed, this model tends to underestimate the true value function, which can potentially result in overly conservative policies as a consequence.
In contrast, the values estimated by \model{} are notably more accurate and more akin to the true values.

\vspace{-5pt}
\paragraph{Dependence of CoWorld to domain similarities.}
We further investigate the dependence of \model{} on domain similarity from the perspectives of different observation spaces and reward spaces. 
We first explore how \model{} performs when we only have source domains with significantly distinct observation spaces from the target domain.
As illustrated in \tabref{tab:low_dim_comp}, the agent receives low-dimensional state inputs in the source domain (Meta-World) and high-dimensional images in the target domain (RoboDesk). 
We can see that \model{} outperforms \textit{Offline DV2} by $13.3\%$ and $34.0\%$ due to the ability to leverage low-dimensional source data effectively. 
Notably, the finetuning method (\textit{DV2 Finetune}) is not applicable in this scenario. 
In \tabref{tab:diff_reward_comp}, we also observe that \model{} benefits from a source domain, even with a significantly different reward signal.
Unlike previous experiments, we use a sparse reward function for the source Meta-World tasks. It is set to $500$ only upon task completion and remains $0$ before that.
The experimental results demonstrate that although excessively sparse rewards can hinder the training process, \model{} still achieves an average performance gain of $6.6\%$ compared to \textit{DV2 Finetune} under the same setting.

\vspace{-5pt}
\paragraph{Comparison to jointly training one world model across domains.}
Notably, \model{} is implemented with separate world models for the source and target domains. 
Alternatively, we can employ a jointly trained world model across various domains for more efficient memory usage.
In \tabref{tab:comp_oneworld}, we compare the results from the original \model{} and ``\textit{Multi-Task DV2}''. 
Multi-Task DV2 involves training DreamerV2 on both offline and online data with a joint world model and separate actor-critic models. 
\model{} consistently performs better.
Intuitively, using separate world models allows the source and target domains to have different physical dynamics, observation spaces, or reward formations, as the scenarios shown in \tabref{tab:low_dim_comp} and \tabref{tab:diff_reward_comp}.

\begin{table}[t]
\caption{Experiments with significantly distinct observation spaces across domains. We use \textbf{\textit{low-dimensional}} state data as inputs for the RL agents in the source domain and \textbf{\textit{high-dimensional}} image observations in the target domain. \textbf{MW} represents Meta-World and \textbf{RD} stands for RoboDesk.}
\label{tab:low_dim_comp}
\setlength\tabcolsep{10pt}
\centering
\footnotesize
\scalebox{1}{
\begin{tabular}{l|cc}
\toprule
Method & MW: \textit{Button Press} $\rightarrow$ RD: \textit{Push Button} & MW:  \textit{Window Close} $\rightarrow$ RD: \textit{Open Slide} \\
\midrule
Offline DV2 & 347 $\pm$ 24 & 156 $\pm$ 46 \\
CoWorld & \textbf{393 $\pm$ 64} & \textbf{209 $\pm$ 43} \\
\bottomrule
\end{tabular}
}
\end{table}

\begin{table}[t]
\caption{Experiments with significantly distinct reward formations across domains. We use \textbf{\textit{sparse}} rewards in the source domain while maintaining the \textbf{\textit{dense}} rewards in the target domain. }
\label{tab:diff_reward_comp}
\setlength\tabcolsep{9pt}
\centering
\footnotesize
\scalebox{1}{
\begin{tabular}{l|cc}
\toprule
Method & MW: \textit{Button Press} $\rightarrow$ RD: \textit{Push Button} & MW:  \textit{Window Close} $\rightarrow$ RD: \textit{Open Slide} \\
\midrule
DV2 Finetune & 314 $\pm$ 51 & 173 $\pm$ 39 \\
CoWorld & \textbf{335 $\pm$ 28} & \textbf{184 $\pm$ 32} \\
\bottomrule
\end{tabular}
}
\vspace{-5pt}
\end{table}

\begin{table}[thbp]
\caption{Comparison to jointly training \textbf{one world model} across domains (\textit{Multi-Task DV2}). }
\label{tab:comp_oneworld}
\setlength\tabcolsep{7pt} 
\centering
\footnotesize
\scalebox{1}{
\begin{tabular}{l|cc}
\toprule
Method & MW: \textit{Button Press} $\rightarrow$ RD: \textit{Push Button} & MW:  \textit{Window Close} $\rightarrow$ RD: \textit{Open Slide} \\
\midrule
Multi-Task DV2 & 342 $\pm$ 29 & 173 $\pm$ 22 \\
CoWorld & \textbf{428 $\pm$ 42} & \textbf{202 $\pm$ 19} \\
\bottomrule
\end{tabular}
}
\end{table}

\begin{figure*}[t]
    \centering
    \includegraphics[width=\linewidth]{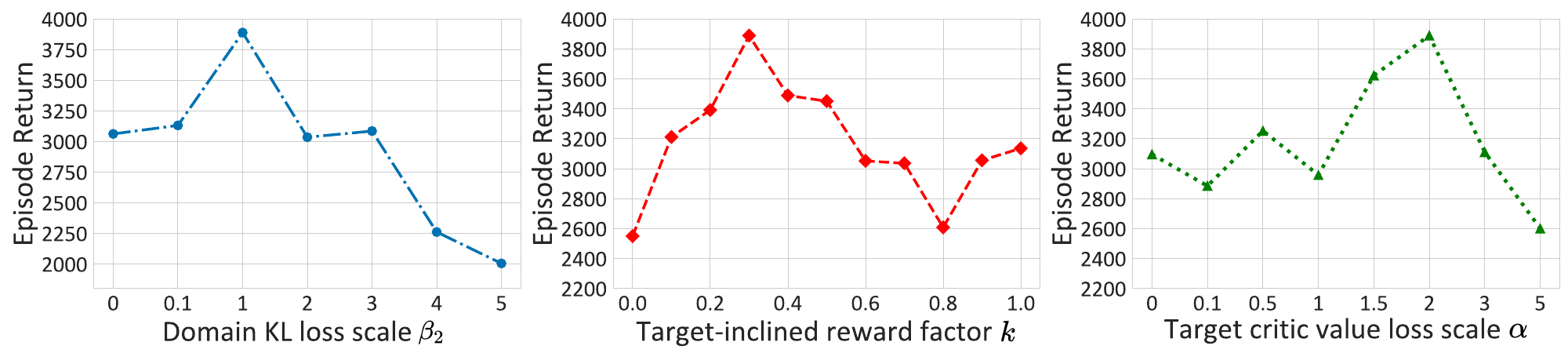}
    \vspace{-15pt}
    \caption{
    Sensitivity analysis of the hyperparameters on Meta-World (\textit{DC} $\rightarrow$ \textit{BP}). 
    }
    \label{fig:sensitivity}
    \vspace{-10pt}
\end{figure*}

\vspace{-5pt}
\paragraph{Hyperparameter sensitivity.} 
We conduct sensitivity analyses on Meta-World (\textit{DC} $\rightarrow$ \textit{BP}).
From \figref{fig:sensitivity}, we observe that when $\beta_2$ for the domain KL loss is too small, the state alignment between the source and target encoders becomes insufficient, hampering the transfer learning process. Conversely, if $\beta_2$ is too large, the target encoder becomes excessively influenced by the source encoder, resulting in a decline in performance.
We also find that the target-informed reward factor $k$ plays a crucial role in balancing the influence of source data and target reward information, which achieves a consistent improvement over \textit{DV2 Finetune} ($2456\pm661$) in the range of $[0.1,0.7]$.
Moreover, we discover that the hyperparameter $\alpha$ for the target value constraint performs well within $[1,3]$, while an excessively larger $\alpha$ may result in value over-conservatism in the target critic.

%% file: sections/05.related.tex
\section{Related Work}
Learning control policies from images is critical in real-world applications. 
Existing approaches can be grouped by the use of model-free~\cite{laskin2020curl,schwarzer2021pretraining,stooke2021decoupling,xiao2022masked,parisi2022unsurprising} or model-based~\cite{hafner2019learning,hafner2020dream,hafner2021mastering,seo2022reinforcement,pan2022iso,hafner2022deep,mazzaglia2023choreographer,micheli2023transformers,zhang2023predictive,ying2023reward} RL algorithms.
In offline RL, agents leverage pre-collected offline data to optimize policies and encounter challenges associated with value overestimation~\cite{levine2020offline}.
Previous methods mainly suggest taking actions that were previously present in the offline dataset or learning conservative value estimations~\citep{fujimoto2019off,kumar2020conservative,chen2022lapo,yu2020mopo,yu2021combo,rigter2022rambo}. 
Recent approaches have introduced specific techniques to address the challenges associated with offline visual RL~\citep{mandlekar2019scaling,dasari2019robonet,levine2020offline,agarwal2020optimistic,rafailov2021offline,yu2022leverage,seo2022reinforcement,zang2023behavior,cho2022s2p,lu2023challenges}.
Rafailov \textit{et al.} \cite{rafailov2021offline} proposed to handle high-dimensional observations with latent dynamics models and uncertainty quantiﬁcation.
Cho \textit{et al.} \cite{cho2022s2p} proposed synthesizing the raw observation data to append the training buffer, aiming to mitigate the issue of overfitting.
In a related study, Lu \textit{et al.} \cite{lu2023challenges} established a competitive offline visual RL model based on DreamerV2~\cite{hafner2021mastering}, so that we use it as a significant baseline of our approach. 
%

Our work is also related to transfer RL, which is known as to utilize the knowledge learned in past tasks to facilitate learning in unseen tasks~\citep{zhu2020transfer,sekar2020planning,zhang2020invariant,sun2021temple,zhang2021learning,eysenbach2021off,yang2021representation,sun2022transfer,ghosh2023reinforcement,kumar2023offline,rafailov2023moto,liu2023cross,nakamoto2023cal}.  
Most existing approaches related to offline dataset + simulator focus on the offline-to-online setup, where the policy is initially pretrained on the offline dataset and then finetuned and deployed on an interactive environment~\cite{nakamoto2023cal, zhang2023policy, yu2023actor, zheng2023adaptive}. These methods aim to bridge the gap between offline and online learning and facilitate fast adaptation of the model to the online environment. In contrast, we explore the online-to-offline setup, which provides a new remedy for the value over-estimation problem.
Additionally, Niu \textit{et al.} \cite{niu2022trust} introduces a dynamics-aware hybrid offline-and-online framework to integrate offline datasets and online simulators for policy optimization. Unlike \model{}, this method primarily focuses on low-dimensional MDPs and cannot be directly used in visual control tasks.
In the context of visual RL, CtrlFormer~\cite{mu2022ctrlformer} learns a transferable state representation via a sample-efficient vision Transformer.
APV~\cite{seo2022reinforcement} executes action-free world model pretraining on source-domain videos and finetunes the model on downstream tasks.
Choreographer~\cite{mazzaglia2023choreographer} builds a model-based agent that exploits its world model to learn and adapt skills in imaginations, the learned skills are adapted to new domains using a meta-controller.
VIP~\cite{ma2023vip} presents a self-supervised, goal-conditioned value-function objective, which enables the use of unlabeled video data for model pertaining.
Unlike previous methods, we handle offline visual RL using auxiliary simulators, mitigating the value overestimation issues with co-trained world models.

%% file: sections/06.conclusion.tex
\section{Conclusions and Limitations}
\label{sec:conclusion}
In this paper, we proposed a transfer RL method named \model{}, which mainly tackles the difficulty in representation learning and value estimation in offline visual RL. 
The key idea is to exploit accessible online environments to train an auxiliary RL agent to offer additional value assessment. 
To address the domain discrepancies and to improve the offline policy, we present specific technical contributions of cross-domain \textit{state alignment}, \textit{reward alignment}, and \textit{min-max value constraint}.
\model{} demonstrates competitive results across three RL benchmarks.
An unsolved problem of \model{} is the increased computational complexity associated with the training phase in auxiliary domains (see \underline{Appendix \ref{sec:training_efficiency}}). It is valuable to improve the training efficiency in future research.

\section*{Acknowledgments}
This work was supported by the National Natural Science Foundation of China (Grant No. 62250062, 62106144, 62302246), the Shanghai Municipal Science and Technology Major Project (No. 2021SHZDZX0102), the Fundamental Research Funds for the Central Universities, and the CCF-Tencent Rhino-Bird Open Research Fund. 
This work was also supported by the Natural Science Foundation of Zhejiang Province, China (No. LQ23F010008), the High-Performance Computing Center at Eastern Institute of Technology (Ningbo), and the Ningbo Institute of Digital Twin.

%% file: sections/07.appendix.tex
\appendix
\section*{Appendix}

In this appendix, we provide the following supplementary materials:
(\ref{sec:apdx_a}) Details of the proposed model, including further descriptions of the learning schemes, the notations, the world model architecture, the behavior learning objective functions, and hyperparameters.
(\ref{sec:more_results}) Additional experimental results, including visualization of the learned policy, quantitative results on offline datasets with mixed data quality, comparison to using a pre-trained foundation model such as R3M, and computational efficiency.
(\ref{sec:multi_source}) Implementation details of the multi-source \model{} model and further empirical analysis on the selected source domain.
(\ref{sec:domains}) Detailed setups of the source and target domains.
(\ref{sec:compared_methods}) Details of the compared methods.
(\ref{sec:borader_impacts}) Potential social impacts of the proposed method.

\section{Model Details}
\label{sec:apdx_a}

\subsection{Framework of \model{}}
\label{appendix_a}

\begin{figure}[h]
    \centering
    \includegraphics[width=\textwidth]{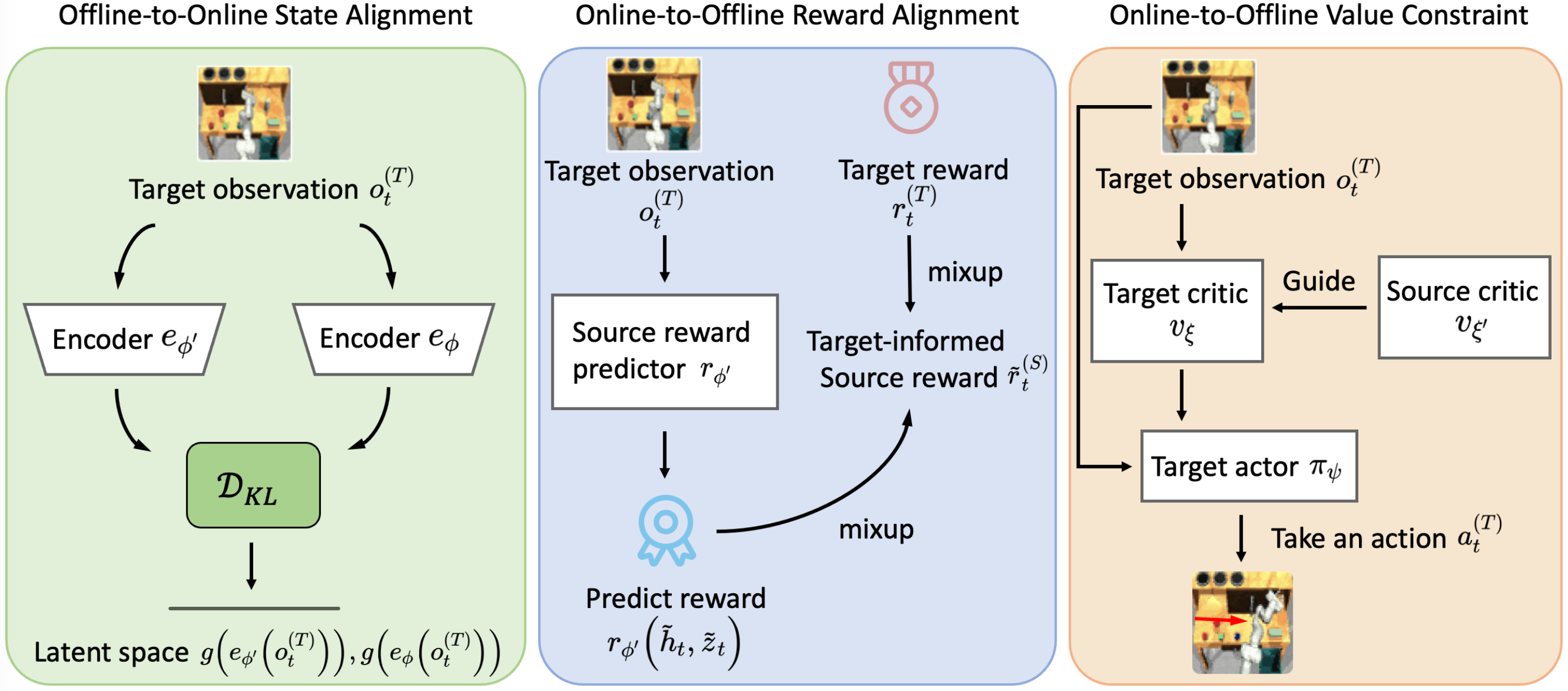}
    \vspace{-15pt}
    \caption{
    \model{} uses an auxiliary online environment to build a policy ``\textit{test bed}'' that is aware of offline domain information. This, in turn, can guide the visual RL agent in the offline domain to learn a mildly-conservative policy, striking a balance between value overestimation and over-conservatism.
     }
    \label{fig:coworld_framework}
    \vspace{-10pt}
\end{figure}

As illustrated in \figref{fig:coworld_framework}, the entire training process of \model{} comprises three iterative stages: offline-to-online state alignment (Stage A), online-to-offline reward alignment (Stage B), and online-to-offline value constraint (Stage C). 
First, we feed the same target domain observations sampled from $\mathcal{B}^{(T)}$ into the encoders and close the distance of $e_{\phi^{\prime}}(o_t^{(T)})$ and $e_{\phi}(o_t^{(T)})$ in Stage A.
Second, in Stage B, the source reward predictor $r_{\phi^{\prime}}(\cdot)$ is trained with mixed data from both of the replay buffers $\mathcal{B}^{(S)}$ and $\mathcal{B}^{(T)}$. Notably, when we sample data from $\mathcal{B}^{(T)}$, the reward will be relabelled as the target-informed source reward.
Finally, we introduce a min-max value constraint using the source critic to the target critic in Stage C.

For notations, we use the superscript $S$ and $T$ to represent data from the source and target domains. Additionally, subscripts $(\phi^\prime, \psi^\prime, \xi^\prime)$ and $(\phi, \psi, \xi)$ are employed to distinguish model parameters for different domains. The notations of source and target domains are summarised in \tabref{tab:notations}.

\begin{table*}[h]
\vspace{-5pt}
\caption{Notations of the source and target domains. } 
\label{tab:notations}
\setlength\tabcolsep{5pt}
\begin{center}
\begin{tabular}{p{2.5cm}|p{4cm}|p{6.4cm}}
\toprule
Domains & Model Parameters & Data  \\
\midrule
Source/Online~($S$) & World model $\phi^\prime$, Actor $\psi^\prime$, Critic $\xi^\prime$ & Raw data $(o_t^{(S)}, a_t^{(S)}, r_t^{(S)})$, Relabeled reward with infomation from both domains $\tilde{r}_t^{(S)}$ \\
Target/Offline~($T$) & World model $\phi$, Actor $\psi$, Critic $\xi$ & Raw data $(o_t^{(T)}, a_t^{(T)}, r_t^{(T)})$ \\
\bottomrule
\end{tabular}
\end{center}
\vspace{-5pt}
\end{table*}

\subsection{World Model}
\label{sec:world_model_learning}
We adopt the framework of the world model used in ~\cite{hafner2021mastering}. 
The image encoder is a Convolutional Neural Network~(CNN). 
The image predictor is a  transposed CNN and the transition, reward, and discount factor predictors are Multi-Layer Perceptrons~(MLPs). 
The discount factor predictor serves as an estimate of the probability that an episode will conclude while learning behavior based on model predictions.
The encoder and decoder take $64 \times 64$ images as inputs.

\subsection{Behavior Learning}
\label{sec:bl}
For the behavior learning of \model{}, we use the actor-critic method from DreamerV2~\citep{hafner2021mastering}.
The $\lambda$-target $V_t^{(T)}$ in \eqref{eq:critic_loss} is defined as follows:
\begin{equation}
V_t^{(T)} \doteq \hat{r}^{(T)}_t+\hat{\gamma}_t^{(T)} \begin{cases}(1-\lambda) v_{\xi}\left(\hat{z}_{t+1}^{(T)}\right)+\lambda V_{t+1}^{(T)} & \text { if } t<H \\ v_{\xi}\left(\hat{z}_H^{(T)}\right) & \text { if } t=H\end{cases},
\end{equation}
where $\lambda$ is set to $0.95$ for considering more on long horizon targets.
The actor and critic are both MLPs with ELU activations~\cite{clevert2015fast}. 
The target actor and critic are trained with guidance from the source critic and regress the $\lambda$-return with a squared loss.
The world model is fixed during behavior learning.
The source actor and critic are:
\begin{equation}
\begin{alignedat}{3}
& \text{Source Actor:}   \padspace && \hat{a}_t^{(S)} \sim \pi_{\psi^{\prime}}(\hat{a}_t^{(S)} | \hat{z}_t^{(S)}) \\
& \text{Source Critic:}  \padspace && v_{\xi^{\prime}}(\hat{z}_t^{(S)}) \approx \mathbb{E}_{p_{\phi^{\prime}},p_{\psi^{\prime}}}\Big[
  \textstyle\sum_{\tau \geq t} \hat{\gamma}_{\tau-t}^{(S)} \hat{r}_\tau^{(S)}
\Big]. \\
\end{alignedat}
\end{equation}

We train the source actor $\pi_{\psi^{\prime}}$ by maximizing 
\begin{equation}
\label{eq:sc_actor_loss}
\begin{aligned}
    &\mathcal{L}(\psi^{\prime})  = \ \mathbb{E}_{p_{\phi^{\prime}}, p_{\psi^{\prime}}} \Big[\sum_{t=1}^{H-1}(
    \underbrace{\beta \mathrm{H} \left[a_t^{(S)} \mid \hat{z}_t^{(S)} \right]}_{\text{entropy regularization}}
    + \underbrace{\rho V_t^{(S)}}_{\text{dynamics backprop}} \\ &+ 
    \underbrace{(1-\rho) \ln \pi_{\psi^{\prime}}(\hat{a}_t^{(S)} \mid \hat{z}_t^{(S)}) \texttt{sg}(V_t^{(S)}-v_{\xi^{\prime}}(\hat{z}_t^{(S)}))}_{\text{REINFORCE}}\Big].
\end{aligned}
\end{equation}

The source critic $v_{\xi^{\prime}}$ is optimized
by minimizing
\begin{equation}
\label{eq:sc_critic_loss}
\begin{aligned}
    \mathcal{L}(\xi^{\prime}) = \mathbb{E}_{p_{\phi^{\prime}}, p_{\psi^{\prime}}}\Big[\sum_{t=1}^{H-1} \frac{1}{2}\left(v_{\xi^{\prime}}\left(\hat{z}_t^{(S)}\right)-\texttt{sg}\left(V_t^{(S)}\right)\right)^2\Big].
\end{aligned}
\end{equation}

\subsection{Hyperparameters}
\label{sec:hyper}
The hyperparameters of \model{} are shown in \tabref{tab:hparams}. 
\begin{table}[htbp]
\centering
\caption{Hyperparameters of \model{}.} 
\vskip 0.05in
\setlength{\tabcolsep}{4mm}{} 
\begin{tabular}{lccc}
\toprule
\textbf{Name} & \textbf{Notation} & \multicolumn{2}{c}{\textbf{Value}} \\
\midrule
\texttt{Co-training} & &\multicolumn{1}{c}{Meta-World / RoboDesk} &\multicolumn{1}{c}{DMC}  \\
\midrule
Domain KL loss scale & $\beta_2$ & $1$ & $1.5$\\
Target-informed reward factor & $k$ & $0.3$ & $0.9$ \\
Target critic value loss scale & $\alpha$ & $2$ &1 \\
Source domain update iterations & $K_1$ & $2\cdot10^{4}$&$2\cdot10^{4}$ \\
Target domain update iterations & $K_2$ & $5\cdot10^{4}$&$2\cdot10^{4}$\\
\midrule
\texttt{World Model} \\
\midrule
Dataset size  & --- &  \multicolumn{2}{c}{$2\cdot10^6$} \\
Batch size & $B$ &  \multicolumn{2}{c}{50} \\
Sequence length & $L$ & \multicolumn{2}{c}{50} \\
KL loss scale & $\beta_1$ & \multicolumn{2}{c}{1} \\
Discrete latent dimensions & --- & \multicolumn{2}{c}{32} \\
Discrete latent classes & --- & \multicolumn{2}{c}{32} \\
RSSM number of units & --- & \multicolumn{2}{c}{600} \\
World model learning rate & --- & \multicolumn{2}{c}{$2\cdot10^{-4}$} \\
\midrule
\texttt{Behavior Learning} \\
\midrule
Imagination horizon & $H$ & \multicolumn{2}{c}{15} \\
Discount & $\gamma$ & \multicolumn{2}{c}{0.995} \\
$\lambda$-target  & $\lambda$ & \multicolumn{2}{c}{0.95} \\
Actor learning rate & --- & \multicolumn{2}{c}{$4\cdot10^{-5}$} \\
Critic learning rate & --- & \multicolumn{2}{c}{$1\cdot10^{-4}$} \\
\bottomrule
\end{tabular}
\label{tab:hparams}
\end{table}

\section{Additional Quantitative and Qualitative Results}
\label{sec:more_results}

\subsection{Visualizations on Policy Evaluation}
\label{sec:vis_policy_evaluation}
\begin{figure}[htbp]
    \centering
    \subfigure[Policy evaluation on the DMC \textit{Walker Downhill} task]{\includegraphics[width=\linewidth]{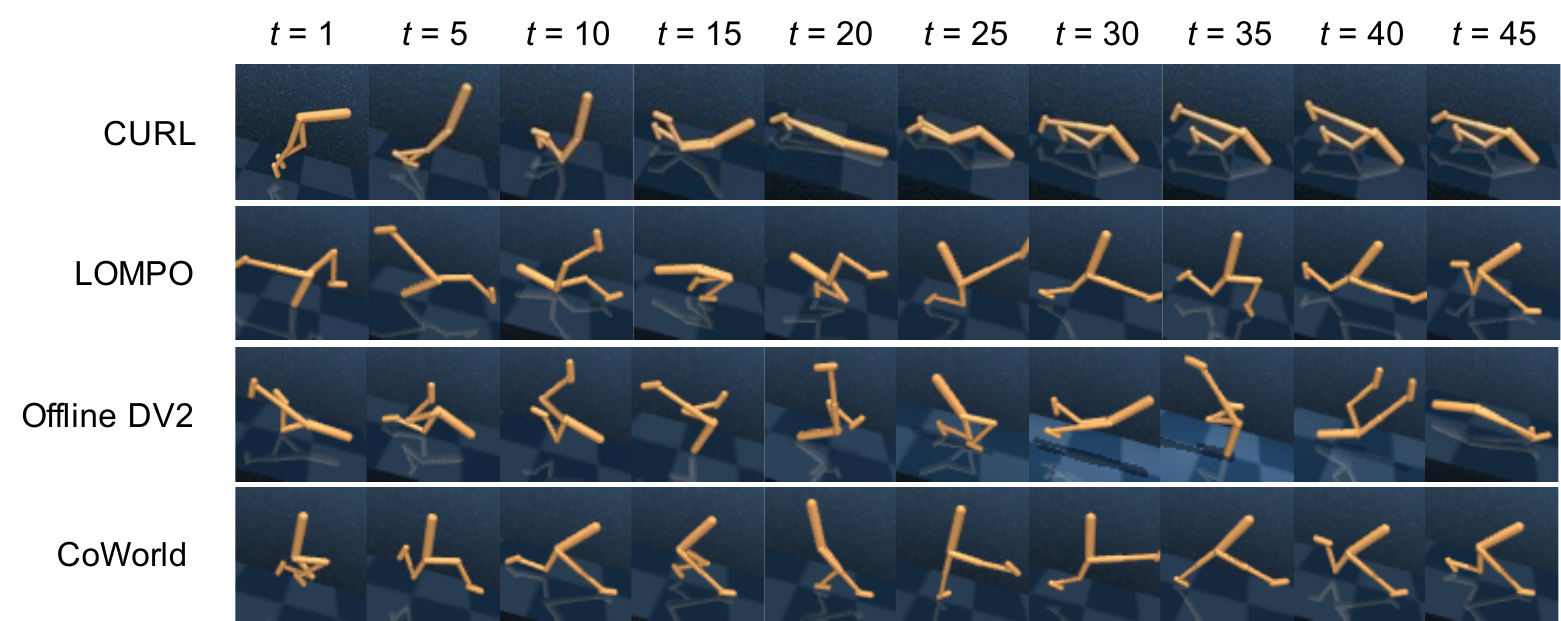} 
    \label{fig:walk_do_com}
    }
    \subfigure[Policy evaluation on the DMC \textit{Walker Uphill} task]{\includegraphics[width=\linewidth]{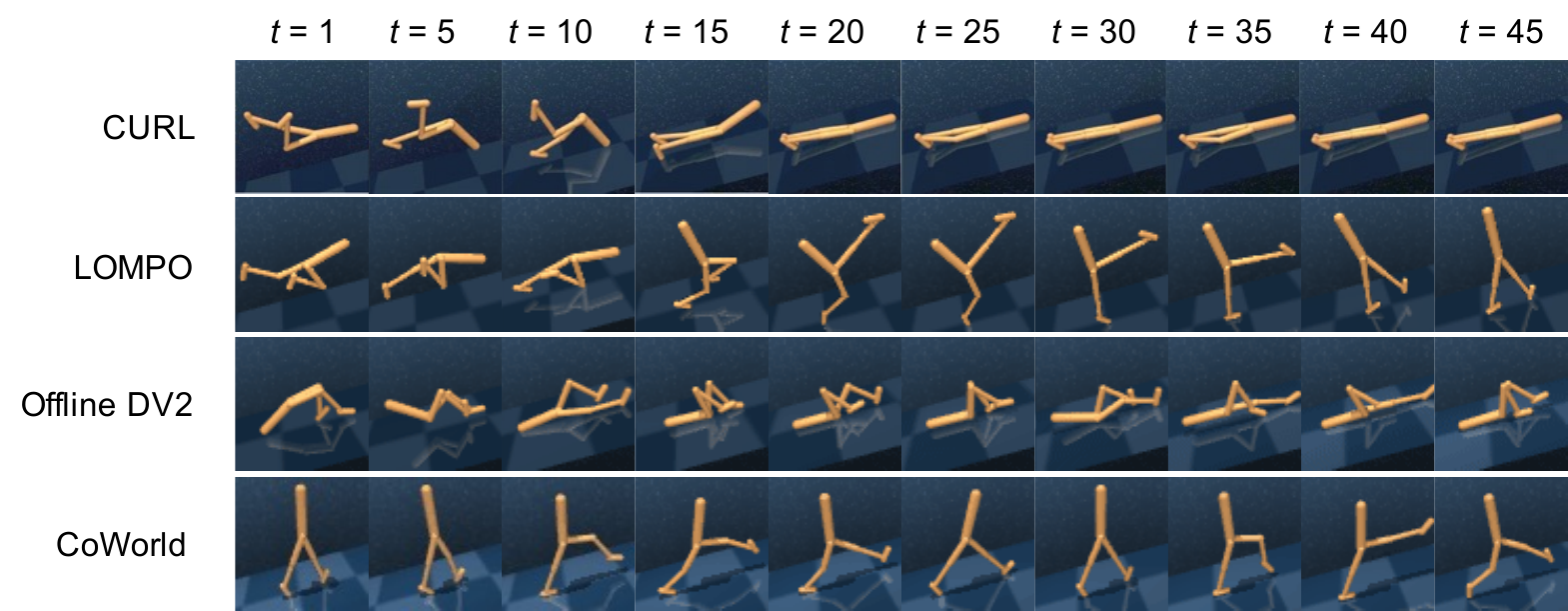}
    \label{fig:walk_up_com}
    }
    \subfigure[Policy evaluation on the DMC \textit{Walker Nofoot} task]{\includegraphics[width=\linewidth]{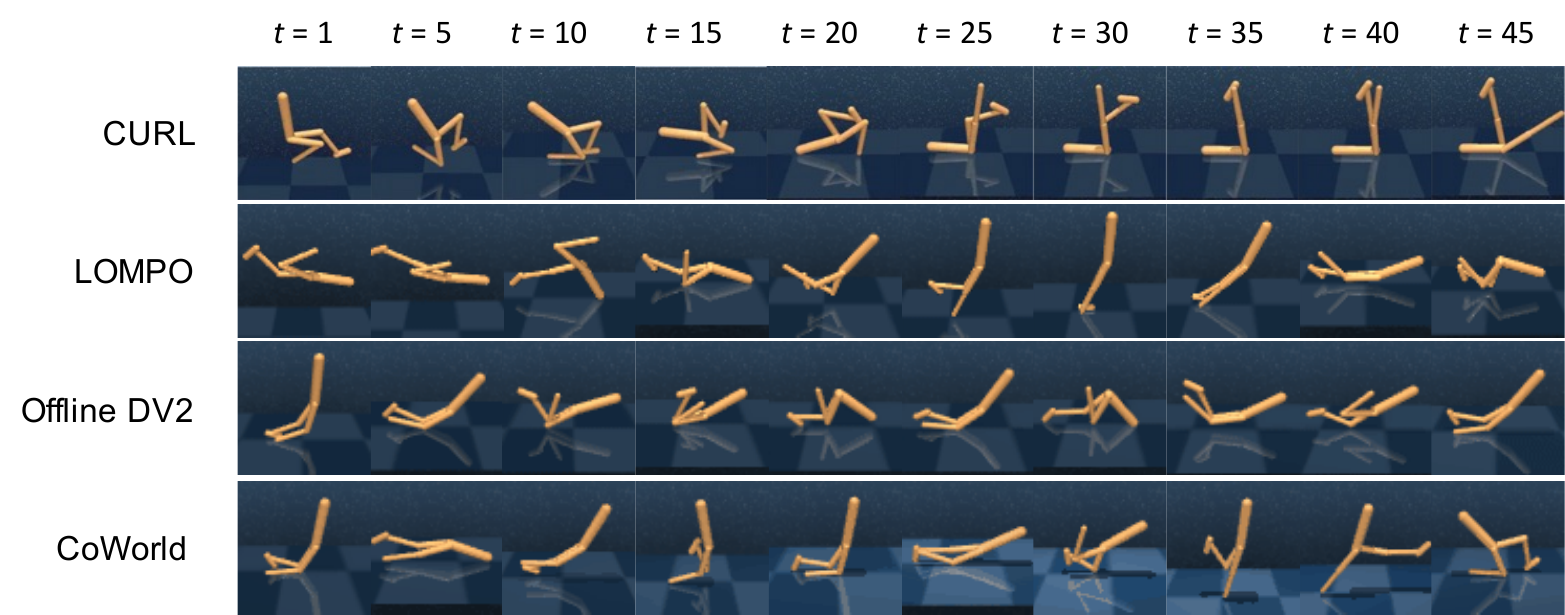}
    \label{fig:walk_no_cmp}
    }
    \vspace{-8pt}
    \caption{
    Additional qualitative results of policy evaluation on the DMC tasks.
    }
    \label{fig:walk_cmp}
\end{figure}

\begin{figure}[t]
    \centering
    \includegraphics[width=\linewidth]{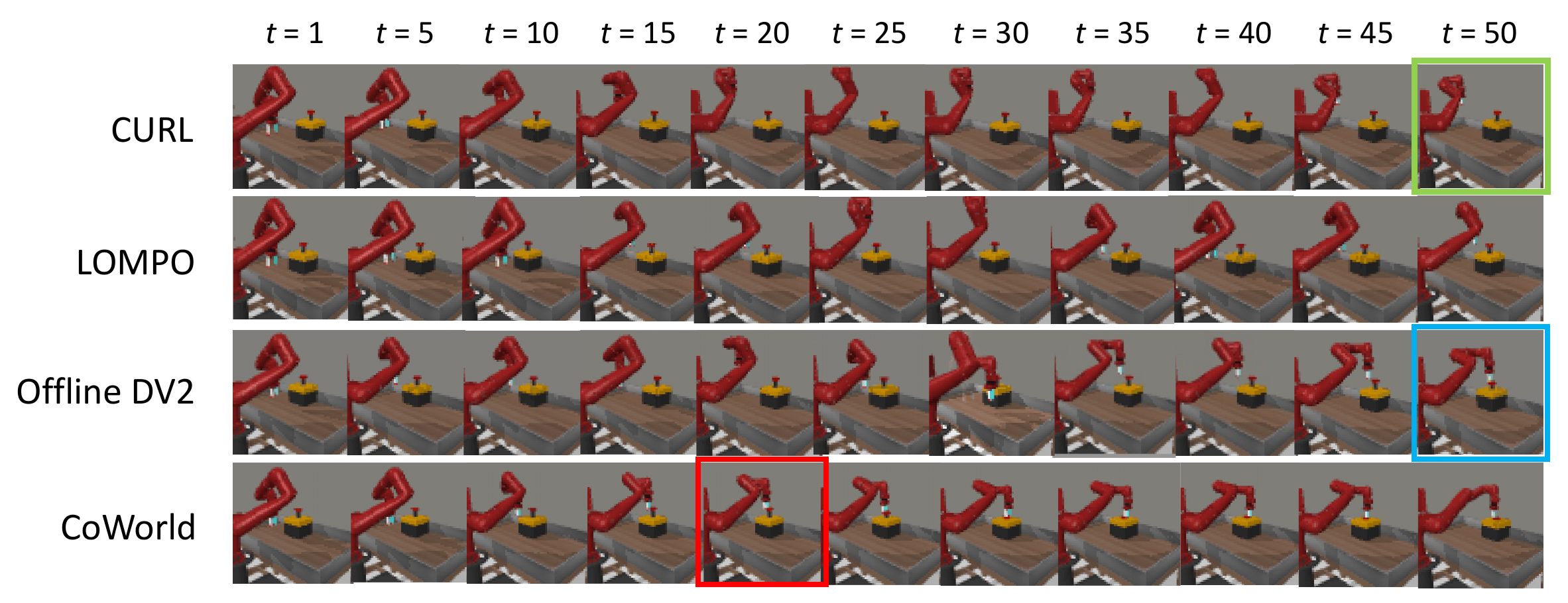}
    \vspace{-15pt}
    \caption{Policy evaluation on the Meta-World \textit{Button Topdown} task. The model-free method \textit{CURL} cannot complete the task (\textcolor{MyDarkGreen}{green box}). \model{} achieves better performance and finishes the task in fewer steps (\textcolor{MyDarkRed}{red box}) than \textit{Offline DV2} (\textcolor{MyDarkBlue}{blue box}).}
    \label{fig:policy_eval_meta}
\end{figure}

We evaluate the trained agent of different models on the Meta-World and DMC tasks and select the first $45$ frames for comparison.
\figref{fig:walk_cmp} and \figref{fig:policy_eval_meta} present examples of performing the learned policies of different models on DMC and Meta-World respectively. 

\subsection{Quantitative Results on DMC \textit{Medium-Expert} Dataset}
\label{sec:dmc_expert_res}
Similar to the data collection strategy of the \textit{medium-replay} dataset, we build offline datasets with \textit{medium-expert} quality using a DreamerV2 agent.  
The \textit{medium-expert} dataset comprises all the samples in the replay buffer during the training process until the policy attains expert-level performance, defined as achieving the maximum score that the DreamerV2 agent can achieve. As shown in \tabref{tab:dmc_expert_result}, CoWorld outperforms other baselines on the DMC \textit{medium-expert} dataset in most tasks.

\begin{table*}[h]
\vspace{-5pt}
\caption{Performance on DMC \textit{medium-expert} dataset. We report the mean rewards and standard deviations of $10$ episodes over $3$ seeds. } 
\label{tab:dmc_expert_result}
\setlength\tabcolsep{3pt}
\begin{center}
\begin{tabular}{l|ccccccc}
\toprule

Model& WW $\rightarrow$ WD & WW $\rightarrow$ WU &  WW $\rightarrow$ WN  & CR $\rightarrow$ CD & CR $\rightarrow$ CU & CR $\rightarrow$ CN & Avg. \\

\midrule
Offline DV2& $450\pm24$ & $141\pm1$ &   $214\pm8$  & $248\pm9$ & 3$\pm0$ & $48\pm3$ & 184\\
DrQ+BC& \underline{808$\pm$47} & \underline{762$\pm$61} &   \underline{808$\pm$45}  & \underline{862$\pm$13} & \underline{454$\pm$12} & 730$\pm$17 &737 \\
LOMPO  &548$\pm$245& 449$\pm$117 &   688$\pm$97  & 174$\pm$29 & 19$\pm$10 & 113$\pm$35 & 332\\
\cmidrule{1-8}
Finetune& 784$\pm$46 & 671$\pm$65 &  851$\pm$91  & 858$\pm$9 & 428$\pm$49 & \underline{833$\pm$7} &\underline{738} \\  
\model{} & \bfseries  848$\pm$9  & \bfseries 774$\pm$29 & \bfseries 919$\pm$7  & \bfseries 871$\pm$13& \bfseries 475$\pm$16& \bfseries  844$\pm$1 & \bfseries 789\\ 
\bottomrule
\end{tabular}
\end{center}
\vspace{-5pt}
\end{table*}

\subsection{Quantitative Results on Meta-World}
\label{sec:quantitative_results_meta_world}
\begin{figure}[htbp]
    \centering
    \subfigure[Comparison with other baselines]{\includegraphics[width=0.46\textwidth]{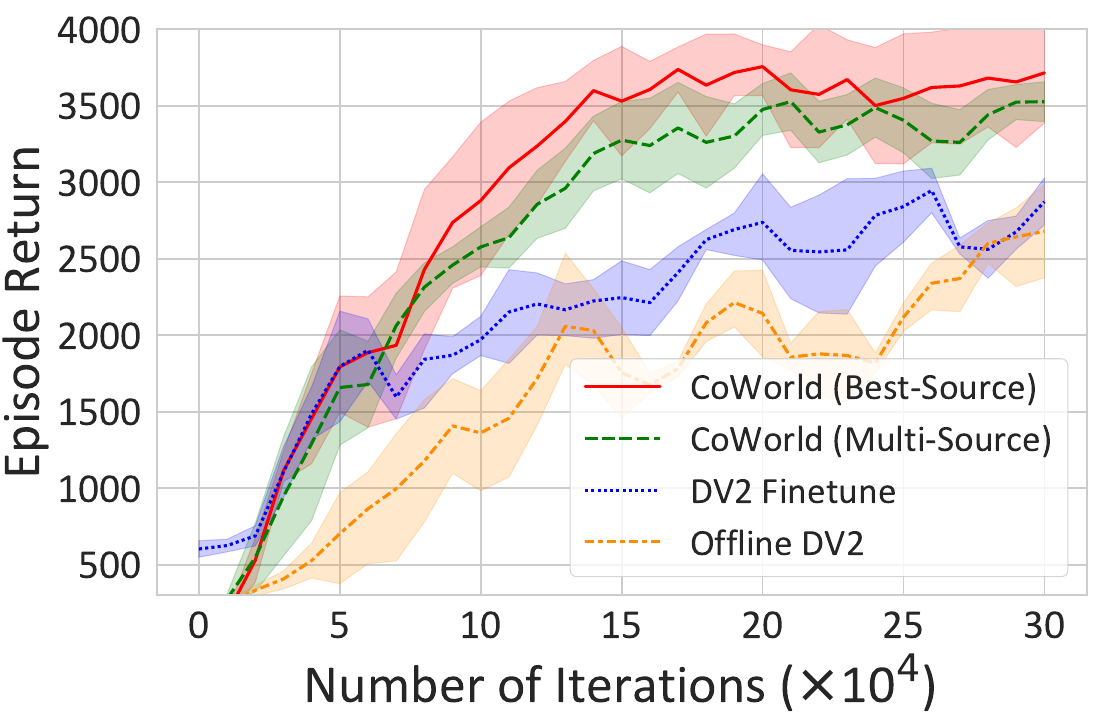} 
    \label{fig:sota_cmp}
  }
    \subfigure[Ablation studies of \model{}]{\includegraphics[width=0.46\textwidth]{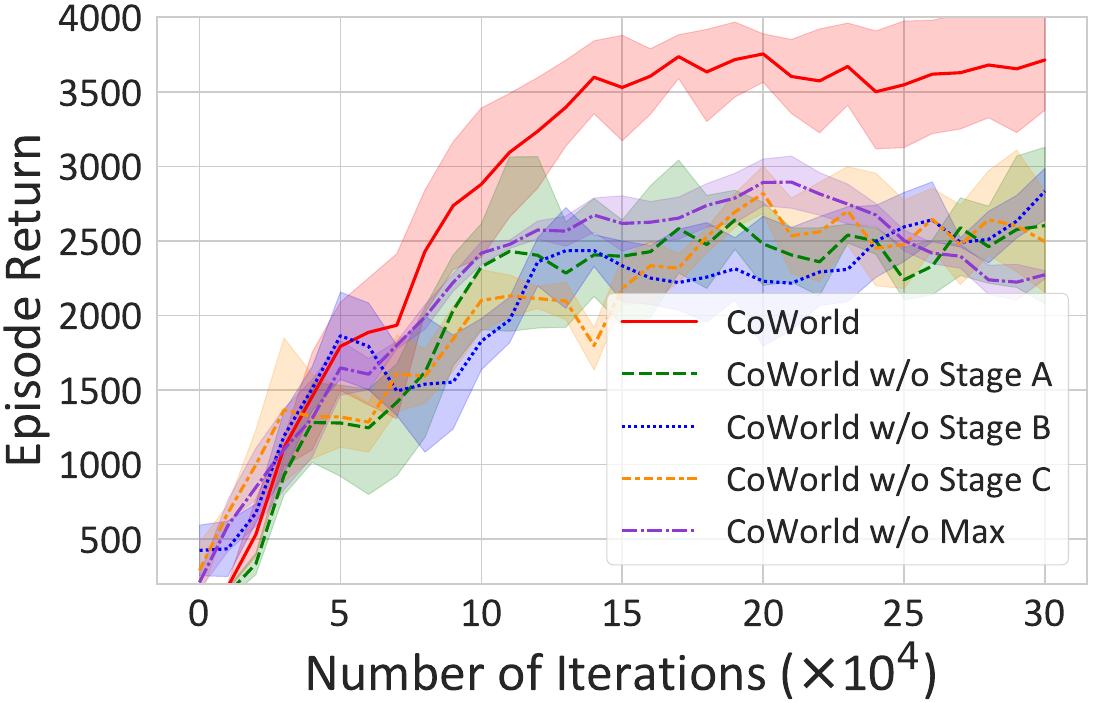}
    \label{fig:metaworld_ablation}
    }
    \vspace{-5pt}
    \caption{
    \textbf{(a)} Comparison with various approaches on the Meta-World \textit{Button Press} task.
    \textbf{(b)} Ablation studies on the Meta-World \textit{Button Press} task that can show the effect of state alignment (\textcolor{MyDarkGreen}{green}), reward alignment (\textcolor{MyPurple}{purple}), and min-max value constraint (\textcolor{MyYellow}{orange}).
    }
    \label{fig:coworld_performance}

\end{figure}

\figref{fig:sota_cmp} compares the performance of different models on Meta-World.
\textit{DV2 Finetune} demonstrates better performance in the initial training phase, thanks to its direct access to the source environment. 
Instead, \model{} introduces auxiliary source value guidance to assist the training of the target agent. 
In the final phase of training, the source value guidance is more effective, and then \model{} outperforms \textit{DV2 Finetune}.  
\figref{fig:metaworld_ablation} presents the ablation studies of \model{} conducted on Meta-World, highlighting the effectiveness and necessity of each training stage.

\subsection{Effect of Latent Space Alignment}
\label{sec:latent_space_align}
\begin{figure}[htbp]
    \centering
    \subfigure[Latent space before alignment]{\includegraphics[width=0.44\textwidth]{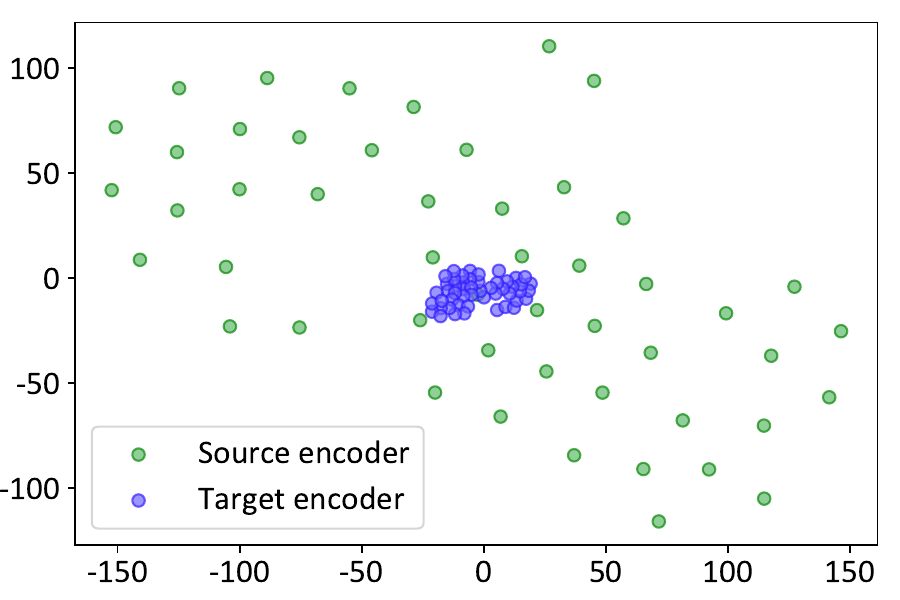} 
    \label{fig:scatter1}
    }
    \subfigure[Latent space after alignment]{\includegraphics[width=0.44\textwidth]{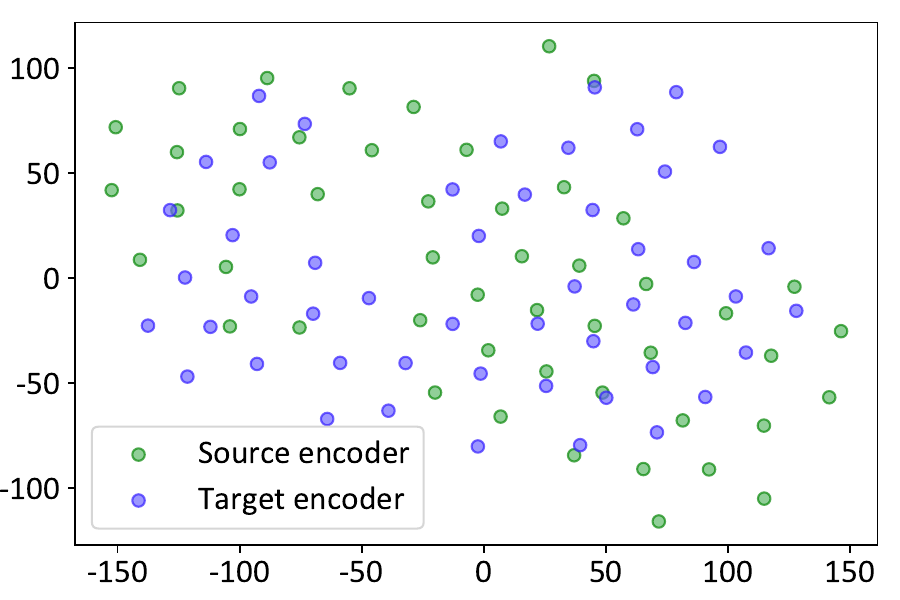}
    \label{fig:scatter2}
    }
    \vspace{-5pt}
    \caption{
    Visualization of the latent space alignment on Meta-World \textit{Handle Press} $\rightarrow$ \textit{Button Press} task by the t-SNE method. (\textbf{a}) Latent space of \model{} before alignment. (\textbf{b}) Latent space of \model{} after alignment.
    }
    \label{fig:multi_scatter}
\end{figure}

We feed the same observations into the source and target encoder of \model{} and then use the t-distributed stochastic neighbor embedding (t-SNE) method to visualize the latent states. 
As shown in \figref{fig:multi_scatter}, the representation learning alignment bridges the gap between the hidden state distributions of the source encoder and target encoder.

\subsection{Additional Results on the Realistic Sim2Real Setup}
Due to the limitation in experimental resources, we are unable to conduct experiments with real robots. We make efforts to construct a more realistic sim2real setup. The experiment is conducted with the identical robot control task for both the source and target domains. We manually introduce two types of noise into the visual observation and action space of the target domain, trying to mimic the complex and noisy real-world scenes.
\begin{itemize}[leftmargin=*]
    \item \textbf{Visual noise}: We modify the original DeepMind Control environment by replacing the static background with dynamic backgrounds of random real-world videos.
    \item \textbf{Action noise}: We add Gaussian noises $n_t$ sampled from $\mathcal{N}(0, 1)$ to every dimension of the action in Meta-World and RoboDesk, which originally ranges in (-1,1). This mimics scenarios where the offline dataset is collected using a low-cost (less inaccurate) real robot.
\end{itemize}
As shown in \tabref{tab:sim2real_result}, we compare \model{} with the finetuned DreamerV2 model on this new setup. We apply noise of three magnitudes, $w \sim \{0.1, 1, 5\}$, in the Meta-World Button Topdown task, leading to noisy actions of $a_t^\text{real} = a_t + w \cdot n_t$.

\begin{table*}[t]
\caption{Results with more significant domain gaps. } 
\label{tab:sim2real_result}
\setlength\tabcolsep{20pt}
\begin{center}
\begin{tabular}{l|cc}
\toprule
Target Domain & CoWorld & DV2 Finetune  \\
\midrule
DMC Walker Walk & \bfseries 544 & 457 \\
DMC Cheetah Run & \bfseries 296 & 220 \\
RoboDesk Push Green ($w=1$) & \bfseries 406 & 358 \\
Meta-World Button Topdown ($w=0.1$) & \bfseries 3752 & 2693 \\
Meta-World Button Topdown ($w=1$) & \bfseries 3567 & 3104 \\
Meta-World Button Topdown ($w=5$) & \bfseries 951 & 670 \\
\bottomrule
\end{tabular}
\end{center}
\vspace{-5pt}
\end{table*}

From the above results, it is evident that CoWorld consistently outperforms the naive finetuning method in this `sim2real' setup. Importantly, we assess the model under more challenging setups, with more significant domain gaps, as illustrated in \tabref{tab:meta_robo_cmp}.

\subsection{Comparison with Pre-trained Foundation Model R3M}
R3M \cite{nair2022r3m} is pretrained on Ego4D human video dataset and facilitates efficient learning of downstream robotic tasks. R3M is shown to be a competitive model, particularly in its ability to transfer representations across domains with diverse visual inputs. For model comparison, we leverage the pre-trained weights of R3M from the official repository to initialize the representation model. We then perform policy optimization based on it for downstream tasks in the Meta-World environment. We respectively employ expert data, sourced from the official repository, alongside our own data, which is collected from scratch with mixed data quality. And \textit{DV2 Finetune} is also pretrained in a related task and finetuned on the offline dataset. As demonstrated in \tabref{tab:compared_r3m}, our approach outperforms the R3M / DV2 fine-tuning model.

\begin{table*}[t]
\caption{Comparison of \model{} with using a pre-trained foundation model, R3M. } 
\label{tab:compared_r3m}
\setlength\tabcolsep{5pt}
\begin{center}
\begin{tabular}{l|cccc}
\toprule
 & \model{} & R3M~(expert data) & R3M~(our data) & DV2 Finetune \\
\midrule
Button Press Topdown & \bfseries 3889 & 1609 & 311 & 3499 \\
Drawer Close & \bfseries 4845 & N/A & 4616 & 4273 \\
Handle Press & \bfseries 4570 & N/A & 1603 & 3702  \\
\bottomrule
\end{tabular}
\end{center}
\vspace{-5pt}
\end{table*}

It is important to note that:
\begin{itemize}[leftmargin=*]
    \item Despite its generalizable representations, R3M is NOT specifically designed to solve the value overestimation problem, which is fundamental in offline RL. In contrast, our approach not only aligns state representations across domains but also effectively tackles the issue of value overestimation, and therefore achieves better performance.
    \item The fine-tuning process of R3M necessitates expert demonstrations for high-quality imitation learning. However, its performance empirically deteriorates when applied to the offline dataset of the \textit{medium-replay} data.
    \item The pre-training process of R3M typically takes around $5$ days on a V100 GPU, while the entire training procedure of our approach takes only about $2$ days on a 3090 GPU.
\end{itemize}

\subsection{Training Efficiency}
\label{sec:training_efficiency}
As shown in \tabref{tab:metaworld_time_complexity}, we evaluate the training/inference time on Meta-World (Handle Press $\rightarrow$ Button Topdown) using a single RTX 3090 GPU. 
Empirically, \model{} achieves convergence ($90\%$ of the highest returns) in approximately $14$ hours; while it costs \textit{DV2 Finetune} about $13$ hours. These results indicate that \model{} requires a comparable training wall-clock time to \textit{DV2 Finetune}, while consistently maintaining better performance in terms of returns after model convergence.

\begin{table*}[t]
\caption{Runtime comparisons evaluated on Meta-World (HP $\rightarrow$ BT). } 
\label{tab:metaworld_time_complexity}
\setlength\tabcolsep{7pt}
\begin{center}
\begin{tabular}{l|ccc}
\toprule

Model& \# Training iterations & Training time&  Inference time per episode  \\

\midrule
Offline DV2 & 300k & 2054 min &2.95 sec \\
DrQ+BC & 300k & 200 min &2.28 sec \\
CQL & 300k & 405 min &1.88 sec \\
CURL & 300k & 434 min &2.99 sec \\
LOMPO & 100k & 1626 min &4.98 sec \\
DV2 Finetune & 460k & 1933 min &6.63 sec \\
DV2 Finetune+EWC& 460k & 1533 min &5.58 sec \\
\model{} & 460k & 3346 min &4.47 sec \\
\bottomrule
\end{tabular}
\end{center}
\vspace{-5pt}
\end{table*}

\section{Multi-Source \model{}}
\label{sec:multi_source}

The key idea of multi-source \model{} is to allocate a set of one-hot weights $\omega_t^{i=1:M}$ to candidate source domains by calculating their KL divergence in the latent state space to the target domain, where $i \in [1, M]$ is the index of each source domain. This procedure includes the following steps: 
\begin{enumerate}[leftmargin=*]
    \item \textbf{World models pretraining:}
    We pretrain a world model for each source domain and target domain individually.
    \item \textbf{Domain distance measurement:}
    At each training step in the target domain, we measure the KL divergence between the latent states of the target domain, produced by $e_{\phi}(o_t^{(T)})$, and corresponding states in each source domain, produced by $e_{\phi^{\prime}_i}(o_t^{(T)})$. Here, $e^{(T)}_{\phi}$ is the encoder of the target world model, and $e_{\phi^{\prime}_i}$ is the encoder of the world model for the source domain $i$.
    \item \textbf{Auxiliary domain identification:} We dynamically identify the closest source domain with the smallest KL divergence. We set $\omega_t^{i=1:M}$ as a one-hot vector, where $\omega_t^i=1$ indicates the selected auxiliary domain.
    \item \textbf{Rest of training:} With the one-hot weights, we continue the rest of the proposed online-to-offline training approach. During representation learning, we adaptively align the target state space to the selected online simulator by rewriting the domain alignment loss term in \eqref{eq:target_world_model_loss} as   
\end{enumerate}
\begin{equation}
\mathcal{L}_{\text{M-S}}=\beta_2 \sum_{i=1}^M \omega_i 
\mathrm{KL}\left[\texttt{sg}(g(e_{\phi^{\prime}}(o_t^{(T)}))) \ \| \ g(e_{\phi}(o_t^{(T)}))\right].
\end{equation}

To evaluate the effectiveness of the multi-source adaptive selection algorithm, we conducted experiments on Meta-World and RoboDesk Benchmark. For each target task, two source tasks are used, including the \model{} best-performing task and the \model{} worst-performing task. Additionally, the sub-optimal source task is added for some target tasks.

As shown in \tabref{tab:multi-source}, multi-source \model{} can adaptively select the best source task for most multi-source problems to ensure adequate knowledge transfer. 
The performance of multi-source \model{} is reported in \tabref{tab:metaworld_result}. 
CoWorld flexibly adapts to the transfer learning scenarios with multiple source domains, achieving comparable results to the model that exclusively uses our manually designated auxiliary simulator as the source domain (best-source). This study significantly improves the applicability of CoWorld in broader scenarios.

\begin{table*}[t]
\caption{The source domain automatically selected by \textit{Multi-Source \model{}}. \textbf{MW} represents Meta-World and \textbf{RD} stands for RoboDesk.} 
\label{tab:multi-source}
\setlength\tabcolsep{30pt}
\begin{center}
\begin{tabular}{l|l}
\toprule
Target domain & Selected source domain  \\
\midrule
MW: Door Close & MW: Drawer Close \\
MW: Button Press & MW: Handle Press \\
MW: Window Close & MW: Button Topdown \\
MW: Handle Press & MW: Button Press \\
MW: Button Topdown & MW: Handle Press \\
MW: Drawer Close & MW: Door Close \\
\cmidrule{1-2}
RD: Push Button & MW: Button Press\\
RD: Open Slide & MW: Window Close\\
RD: Drawer Open & MW: Drawer Close\\
RD: Upright Block off Table & MW: 
 Handle Press\\
\bottomrule
\end{tabular}
\end{center}
\vspace{-5pt}
\end{table*}

\section{Source and Target Domains}
\label{sec:domains}

\paragraph{Meta-World.} 
For the Meta-World environment, we adopt robotic control tasks with complex visual dynamics.
For instance, the \textit{Door Close} task requires the agent to close a door with a revolving joint while randomizing the door positions, and the \textit{Handle Press} task involves pressing a handle down while randomizing the handle positions. 
To evaluate the performance of \model{} on these tasks, we compare it with several baselines in six visual RL transfer tasks.

\vspace{-5pt}
\paragraph{RoboDesk.} 
We select Meta-World as the source domain and RoboDesk as the target domain. Notably, there exists a significant domain gap between these two environments.
The visual observations, physical dynamics, and action spaces of the two environments are different. 
First, Meta-World adopts a side viewpoint, while RoboDesk uses a top viewpoint.
Further, the action space of Meta-World is $4$ dimensional, while that in RoboDesk is $5$-dimensional. 
Considering these differences, the Meta-World $\rightarrow$ RoboDesk benchmark presents a challenging transfer learning problem.

\vspace{-5pt}
\paragraph{DeepMind Control.} 
We train the source agents in standard DMC environments and train the target agents in modified DMC environments. \textit{Walker Uphill} and \textit{Cheetah Uphill} represent tasks in which the ground has a $15^{\circ}$ uphill slope. \textit{Walker Downhill} and \textit{Cheetah Downhill} represent the tasks in which the plane has a $15^{\circ}$ downhill slope.
We evaluate the model in six tasks with different source domains and target domains. 

We assume that there exist notable distinctions between the source and target domains (see \tabref{tab:meta_robo_cmp}).
This assumption can be softened by our proposed approach that mitigates domain discrepancies between distinct source and target MDPs.
Our experiments reveal that the \model{} method exhibits a notable tolerance to inter-domain differences in visual observation, physical dynamics, reward definition, or even the action space of the robots. This characteristic makes it more convenient to choose an auxiliary simulator based on the type of robot. For example: 
\begin{itemize}[leftmargin=*]
\item When the target domain involves a robotic arm (\textit{e.g.}, RoboDesk), an existing robotic arm simulation environment (\textit{e.g.}, Meta-World) can be leveraged as the source domain. 
\item In scenarios with legged robots, environments like DeepMind Control with Humanoid tasks can serve as suitable auxiliary simulators. 
\item For target domains related to autonomous driving, simulators like CARLA can be used.
\end{itemize}

\section{Compared Methods}
\label{sec:compared_methods}

We compare \model{} with several widely used model-based and model-free offline methods.
\begin{itemize}[leftmargin=*]
    \item \textbf{Offline DV2}~\citep{lu2023challenges}:
    A model-based RL method that modifies DreamerV2~\citep{hafner2021mastering} to offline setting, and adds a reward penalty corresponding to the mean disagreement of the dynamics ensemble.
    \item \textbf{DrQ+BC}~\citep{lu2023challenges}:
    It modifies the policy loss term in DrQ-v2~\cite{yarats2021mastering} to match the loss given in \cite{fujimoto2021minimalist}. 
    \item \textbf{CQL}~\citep{lu2023challenges}: 
    It is a framework for offline RL that learns a Q-function that guarantees a lower bound for the expected policy value than the actual policy value. We add the CQL regularizers to the Q-function update of DrQ-v2 \citep{kumar2020conservative}.
    \item \textbf{CURL}~\citep{laskin2020curl}: 
    It is a model-free RL approach that extracts high-level features from raw pixels utilizing contrastive learning.
   \item \textbf{LOMPO}~\citep{rafailov2021offline}: An offline model-based RL algorithm that handles high-dimensional observations with latent dynamics models and uncertainty quantification.
   \item \textbf{LOMPO Finetune}: It pretrains a LOMPO agent~\citep{rafailov2021offline} with source domain data and subsequently finetunes the pretrained agent in the offline target domain.
    \item \textbf{DV2 Finetune}: It pretrains a DreamerV2 agent~\citep{hafner2021mastering} in the online source domain and subsequently finetunes the pretrained agent in the offline target domain. Notably, Meta-World $\rightarrow$ RoboDesk tasks' action space is inconsistent, and we can't finetune directly. Instead, we use the maximum action space of both environments as the shared policy output dimension. For Meta-World and Meta-World $\rightarrow$ RoboDesk transfer tasks, we pretrain the agent for $160$k steps and finetune it $300$k steps. For DMC transfer tasks, we pretrain the agent for $600$k steps and finetune it for $600$k steps. 
    \item \textbf{DV2 Finetune+EWC}: It modifies the \textit{DV2 Finetune} method with EWC~\citep{kirkpatrick2017overcoming} to regularize the model for retaining knowledge from the online source domain. The steps of pretraining and finetuning are consistent with \textit{DV2 Finetune}.
\end{itemize}

\section{Broader Impacts}
\label{sec:borader_impacts}
\model{} is a transfer learning method that may benefit future research in the field of offline RL, model-based RL, and visual RL. 
Beyond the realm of reinforcement learning, this approach holds great potential to contribute to various domains such as robotics and autonomous driving.

In real-world scenarios of healthcare applications, Zhang \textit{et al.} \cite{zhang2023continuous} employed offline RL algorithms to train policies using a large amount of historical dataset, determining the follow-up schedules and tacrolimus dosages in Kidney Transplantation and HIV. There are also corresponding simulators \cite{hua2022personalized, adams2004dynamic} designed by medical domain experts, with parameters learned from real-world data.

Another practical use of the proposed setup is advertising bidding, where direct interactions with real online advertising systems for training are challenging. A recent solution involves constructing a simulated bidding environment based on historical bidding logs for interactive training, such as \cite{mou2022sustainable}, and mitigating the inherent differences between the virtual advertising environment and real-world advertising systems. Therefore, in many real-world scenarios, it is possible to optimize the policies learned from offline datasets with simulators.

A potential negative social impact of our method is the introduction of existing biases from the additional domain. If the training data used to develop our algorithm contains biases, the model may learn those biases, leading to unfair outcomes in decision-making processes. It's crucial to carefully address biases in both data and algorithmic design to mitigate these negative social impacts.